%% file: 0_main.tex
\documentclass[11pt]{article}
\usepackage[final]{colm2026_conference}
\usepackage{microtype}
\usepackage{hyperref}
\usepackage{url}
\usepackage{booktabs}
\usepackage{tabularx}
\usepackage{titletoc}

\usepackage{lineno}

\definecolor{darkblue}{rgb}{0, 0, 0.5}
\hypersetup{colorlinks=true, citecolor=darkblue, linkcolor=darkblue, urlcolor=darkblue}

\usepackage{latexsym}
\usepackage{booktabs}

\usepackage[T1]{fontenc}

\usepackage[utf8]{inputenc}

\usepackage{microtype}

\usepackage{inconsolata}

\usepackage{graphicx}

\usepackage{hyperref}
\usepackage{url}
\usepackage{graphicx}
\usepackage{booktabs}
\usepackage{multirow, multicol}
\usepackage{diagbox}

\usepackage{ctable}
\usepackage{tabularx}
\usepackage{xspace}
\usepackage{amsmath}
\usepackage{float}
\usepackage{xcolor}
\usepackage{listings}
\usepackage{soul}
\usepackage{tablefootnote}
\usepackage{colortbl}

\usepackage{pifont}
\usepackage{amssymb}

\usepackage{enumitem}

\usepackage{listings}
\usepackage{listings-ext}
\usepackage{longtable}
\usepackage{placeins}
\usepackage{textcomp}
\usepackage{fontawesome5}

\usepackage{placeins}
\usepackage{subcaption}
\usepackage{float}

\definecolor{mediumgreen}{RGB}{60, 179, 113}
\definecolor{darkgreen}{rgb}{0.0, 0.5, 0.0}

\lstdefinelanguage{Jinja2}{
  morekeywords={},
  sensitive=false,
  moredelim=[s][\color{blue}]{\{}{\}},
  moredelim=[s][\color{blue}]{\%}{\%},
  moredelim=[s][\color{mediumgreen}]{\{####}{####\}},
}

\lstdefinelanguage{ReAct}{
 basicstyle=\ttfamily\footnotesize,
  morekeywords={},
  sensitive=false,
  morecomment=[l][\color{darkgreen}\bfseries]{Thought:},
  emph={
    >>, search_wikidata, execute_sparql, get_wikidata_entry, stop
  },
  emphstyle={\color{blue}\bfseries},
  stringstyle=\color{red},
  morestring=[b]",
  morestring=[b]""",
    backgroundcolor=\color{lightgray!5},
    captionpos=b,
    tabsize=2,
    xleftmargin=2.0em,
    xrightmargin=2.0em,
    framexleftmargin=2.0em,
    framexrightmargin=2.0em,
    framextopmargin=1em,
    framexbottommargin=1em,
}

\definecolor{comment-red}{rgb}{0.8,0,0}
\definecolor{lightgray}{gray}{0.7}

\definecolor{revcolor}{RGB}{0,0,0}
\newcommand{\rev}[1]{{\color{revcolor}#1}}

\newcommand{\system}{DataSTORM\xspace}

\title{DataSTORM: Deep Research on Large-Scale Databases using Exploratory Data Analysis and Data Storytelling}

\author{Shicheng Liu \quad Yucheng Jiang \quad Sajid Farook \quad Camila Nicollier Sanchez \\\textbf{David Fernando Castro Pena} \quad \textbf{Monica S. Lam} \\
Computer Science Department, Stanford University \\
Stanford, CA \\
\texttt{shicheng, lam@cs.stanford.edu}
}

\begin{document}

\ifcolmsubmission
\linenumbers
\fi
\maketitle

\begin{center}
\small\faGithub\enspace\url{https://github.com/stanford-oval/DataSTORM}
\end{center}
\vspace{-0.5em}

\renewcommand{\thefootnote}{\fnsymbol{footnote}}

\renewcommand{\thefootnote}{\S}
\renewcommand{\thefootnote}{\arabic{footnote}}

\begin{abstract}
Deep research with Large Language Model (LLM) agents is emerging as a powerful paradigm for multi-step information discovery, synthesis, and analysis. However, existing approaches primarily focus on unstructured web data, while the challenges of conducting deep research over large-scale structured databases remain relatively underexplored. Unlike web-based research, effective data-centric research requires more than retrieval and summarization and demands iterative hypothesis generation, quantitative reasoning over structured schemas, and convergence toward a coherent analytical narrative.

In this paper, we present DataSTORM, an LLM-based agentic system capable of autonomously conducting research across both large-scale structured databases and internet sources. Grounded in principles from Exploratory Data Analysis and Data Storytelling, DataSTORM reframes deep research over structured data as a thesis-driven analytical process: discovering candidate theses from data, validating them through iterative cross-source investigation, and developing them into coherent analytical narratives. We evaluate DataSTORM on InsightBench, where it achieves a new state-of-the-art result with a 19.4\% relative improvement in insight-level recall and 7.2\% in summary-level score. We further introduce a new dataset built on ACLED, a large real-world complex database on international conflicts, and demonstrate that DataSTORM outperforms proprietary systems such as ChatGPT Deep Research across both automated metrics and human evaluations.

\end{abstract}

\input{1_introduction}
\input{2_related_work}

\input{3_datastorm}

\input{4_experiments}

\input{6_conclusion}
\input{8_limitations_ethical_considerations}

\bibliography{anthology, custom}
\bibliographystyle{colm2026_conference}

\newpage
\appendix

\input{99_appendix}

\end{document}

%% file: 1_introduction.tex
\section{Introduction}

In many high-value domains, such as finance, scientific experimentation, and policy analysis, the ability to autonomously explore large knowledge sources, synthesize evidence across multiple steps, and converge on meaningful insights is of critical importance. Recently, the NLP community has termed this class of problems "deep research," giving rise to a new generation of agents that combine retrieval, synthesis, and multi-step reasoning over large corpora~\citep{shao-etal-2024-assisting, jiang-etal-2024-unknown, xi2025omnithinkexpandingknowledgeboundaries, li2025webthinkerempoweringlargereasoning, shao2025dr}. However, these systems are overwhelmingly designed for unstructured web text, implicitly assuming that knowledge is primarily expressed in natural language. In many of the domains where deep research matters most, critical knowledge resides not in text, but in structured databases, where it is encoded through numerical relationships, entities, and formal schemas. Despite its importance, deep research over structured data remains largely underexplored~\citep{pérez2025llmbasedapproachinsightgeneration,DBLP:conf/iclr/SahuPRACDTZLVCP25, liu2026huntinsteadwaitevaluating}. As we show in our experiments, even leading systems such as OpenAI Deep Research derive only 23.3\% of their extracted insights from database contents. Conducting deep research in such settings requires fundamentally different capabilities: formulating executable queries, reasoning over quantitative patterns, and maintaining analytical consistency across iterative exploration.

In this work, we argue that enabling deep research over structured data requires rethinking research agents as \textbf{data-driven analytical systems}, rather than purely retrieval-based pipelines. Drawing inspiration from Exploratory Data Analysis (EDA)~\citep{tukey1977exploratory, kirk2016data, wickham2017r} and Data Storytelling~\citep{segel2010narrativevisualization,dykes2019effectivedatastorytelling}, we propose that effective data-centric research must integrate three key properties: (1) iterative hypothesis generation and testing, (2) the interplay of inductive pattern discovery and deductive reasoning, and (3) convergence toward a coherent central narrative.

\begin{figure}[!t]
  \centering
  \includegraphics[scale=0.55]{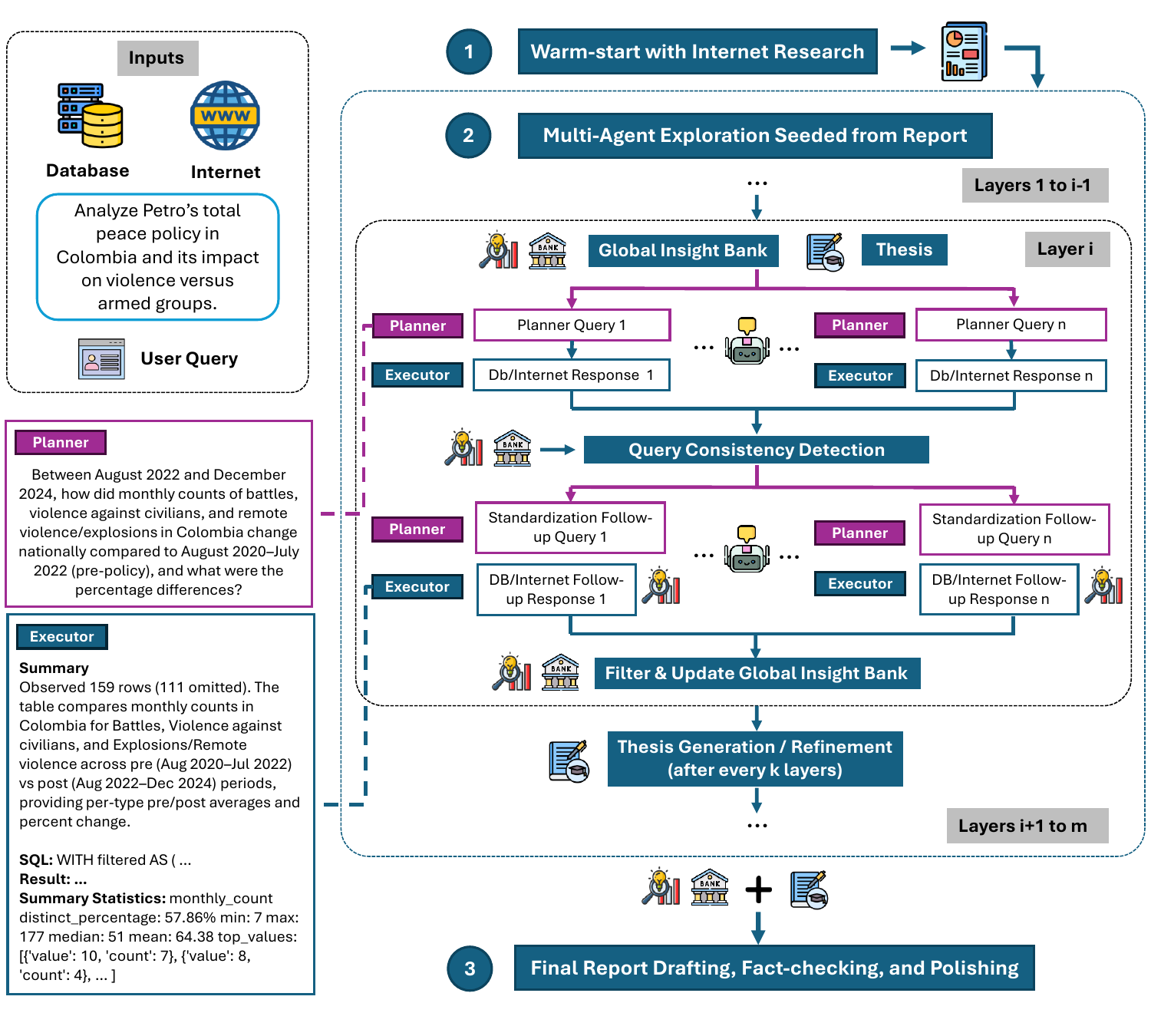}
  \caption{Overview of the DataSTORM system. A complete research workflow consists of three stages: (1) a warm-start module powered by internet research, (2) a multi-agent exploration module, and (3) final report generation.}
  \label{fig:figure1}
\end{figure}

To this end, we introduce DataSTORM, an AI research agent designed to extend deep research beyond text to encompass large-scale structured databases and internet data. DataSTORM introduces a multi-stage workflow (Figure~\ref{fig:figure1}) that combines open-ended exploration with structured analytical reasoning. Our approach makes the following contributions:

\textbf{Multi-agent framework with planner–executor decomposition} \textit{(property 1: iterative hypothesis generation and testing)}: DataSTORM organizes the vast search space of large-scale databases using a hierarchical structure that guides an LLM's exploratory process. As shown in Figure~\ref{fig:figure1}, a high-level \textit{Planner} agent formulates exploratory questions while an \textit{Executor} agent generates SQL queries to retrieve relevant data within an isolated context. This design is rooted in the core EDA cycle of \citet{wickham2017r}: separating the generation of questions (\textit{Planner}) from the search for answers through data retrieval (\textit{Executor}), with each iteration informing the next round of exploration.

\textbf{Integration of inductive statistical discovery with deductive LLM reasoning} \textit{(property 2: interplay of inductive and deductive reasoning)}: Beyond generating SQL queries, the \textit{Executor} proactively surfaces statistical patterns that top-down query strategies alone would not anticipate. Rather than relying solely on LLM-initiated queries, the \textit{Executor} embeds summary statistics directly into its responses, enabling the \textit{Planner} to identify salient patterns such as maximum, minimum, and most frequent values. This synergy between deductive planner-driven exploration and inductive statistical surfacing, an important concept in EDA \citep{kirk2016data}, allows DataSTORM to efficiently uncover non-obvious insights that are otherwise invisible in large-scale databases.

\textbf{Query consistency detection} \textit{(property 3: local analytical coherence toward a central narrative)}: The hierarchical multi-agent structure introduces a challenge that purely retrieval-based pipelines do not face:  independently pursued branches can silently diverge in their analytical criteria (such as time windows and filters), producing findings that appear contradictory but are in fact incomparable. DataSTORM introduces a query consistency detection module that compares each \textit{Executor}'s SQL queries against others at the same level and against the global insight bank, enforcing aligned analytical criteria and ensuring that divergence between findings is meaningful rather than an artifact of inconsistent setup.

\textbf{Thesis-driven exploration} \textit{(property 3: global narrative convergence)}: consistency detections ensure local coherence with respect to formal queries, but do not guarantee that exploration converges on a meaningful narrative. Building on \citet{dykes2019effectivedatastorytelling}'s principle that every data story requires a central insight, DataSTORM introduces a thesis generation module that produces a candidate thesis after an initial phase of exploration, which is continuously refined as further evidence accumulates, transforming open-ended data exploration into a structured progression from discovery to explanation.

We validate DataSTORM on both existing benchmarks and a new benchmark constructed from real-world databases. On InsightBench~\citep{DBLP:conf/iclr/SahuPRACDTZLVCP25}, DataSTORM achieves a new state-of-the-art with a 19.4\% improvement in insight-level scores and 7.2\% in summary-level scores. As existing benchmarks do not fully capture the complexity of deep research over heterogeneous internet and database sources, we construct a new dataset using the Armed Conflict Location \& Event Data (ACLED)~\citep{doi:10.1177/0022343310378914}, a conflict monitoring database maintained by a non-profit organization. On this dataset, DataSTORM outperforms OpenAI Deep Research~\citep{openai2025deepresearch} on both automatic and expert evaluations.

%% file: 2_related_work.tex
\section{Related Work}

\subsection{Agents for deep research and data analysis}

Agents for deep research have become an increasingly active area of research. Most existing systems focus on unstructured internet sources, using retrieval and synthesis over web documents~\citep{shao-etal-2024-assisting, jiang-etal-2024-unknown, xi2025omnithinkexpandingknowledgeboundaries, li2025webthinkerempoweringlargereasoning, shao2025dr}. These systems excel at synthesis of web data but are ill-equipped to reason over structured, quantitative databases, where insights emerge from formal queries and statistical patterns rather than natural language retrieval. DataSTORM addresses this gap by extending deep research to large-scale structured databases.

For agents that focus on data analysis and exploration tasks, many existing works use a fixed pipeline. \citet{ma-etal-2023-insightpilot} define four ``analysis actions'' and use a fixed pipeline to make calls to these actions.
Similarly, \citet{pérez2025llmbasedapproachinsightgeneration}, \citet{yi2025tablepilotrecommendinghumanpreferredtabular}, and \citet{10.1145/3555041.3589724} all utilize a fixed-stage approach, without dynamically adapting their exploration based on intermediate observations. That is, rather than letting each step inform the next, these systems follow a predetermined sequence of modules regardless of what the data reveals. More recent works adopt ReAct-style or multi-agent architectures~\citep{hong2025data, rasheed2024largelanguagemodelsserve, DBLP:conf/iclr/SahuPRACDTZLVCP25, zhang2025deepanalyzeagenticlargelanguage} that allow for more flexible reasoning. However, these approaches rely on general LLM capabilities without specifically addressing challenges unique to large-scale database explorations.

\subsection{Benchmarks for data science agents}

Many benchmarks aim to evaluate the full lifecycle of a data scientist’s workflow. Most of these benchmarks focus on assessing a system’s ability to answer specific data science–oriented questions, similar to traditional text2sql settings~\citep{gu-etal-2024-blade, liu-etal-2024-llms, pmlr-v235-hu24s, majumder2025discoverybench, zhang-etal-2024-benchmarking-data, wang2025fdabenchbenchmarkdataagents, dutta2025condabenchinteractiveevaluationlanguage} as opposed to open-ended research. Others emphasize the construction of end-to-end machine learning pipelines~\citep{10.5555/3692070.3692738}, while some incorporate additional subtasks such as data collection and preprocessing (e.g., reading files and transforming them into the required formats)~\citep{zhang2025datascibenchllmagentbenchmark}.

Overall, benchmarks that evaluate agents in open-ended, database-centric exploration over heterogeneous sources remain limited~\citep{majumder2025discoverybench,DBLP:conf/iclr/SahuPRACDTZLVCP25}. Among existing benchmarks, InsightBench~\citep{DBLP:conf/iclr/SahuPRACDTZLVCP25} most closely evaluates open-ended insight generation for structured databases, and we include it as an evaluation target. To address the remaining gap, we construct a new benchmark using the Armed Conflict Location \& Event Data (ACLED)~\citep{doi:10.1177/0022343310378914}, allowing evaluation against a richer, real-world research scenario.

%% file: 3_datastorm.tex
\section{The \system Framework}

Given as input a relational database $\mathcal{D}$, internet corpus $\mathcal{I}$, and a user query $q$, \system produces a final research report $r$ in free-text form. As shown in Figure~\ref{fig:figure1}, \system consists of three main components: (1) a warm-start phase that produces a lightweight research report $r_0$ based on internet evidence; (2) a multi-agent exploration framework seeded from $r_0$ that runs for a configurable maximum of $m$ layers, producing a final insight bank $B_m = \{b_1, \cdots, b_k\}$ of size $k$, along with a final thesis $t_m$; and (3) a report generation module that drafts, fact-checks, and polishes the final report $r$ based on $B_m$ and $t_m$.

\subsection{Warm-Start with Internet Research}

The warm-start stage provides an initial scaffold for downstream exploration. At this stage, the goal is not to perform exhaustive \textit{deep} internet research, but rather to obtain broad topical coverage and surface promising directions for further investigation. To this end, we employ Co-STORM~\citep{jiang-etal-2024-unknown}, an information-seeking system where multiple language-model agents explore a topic through discourse over retrieved web evidence and synthesize the discovered information into a structured report. Concretely, Co-STORM takes $\mathcal{I}$ as input and produces a list of insights $B_0 =\{b_1, \cdots, b_l\}$ along with a preliminary report $r_0$. These insights are stored in the global insight bank $B_0$. We set starting thesis $t_0$ to be null.

\subsection{Multi-Agent Exploration Framework}
\label{sec:multi-agent-exploration-framework}

\system organizes the exploration space into layers, up to a configurable maximum of $m$ layers. Each layer $i$ takes the current global insight bank $B_{i-1}$ and thesis $t_{i-1}$ as input, and produces an updated insight bank $B_i$.

\subsubsection{Layer Mechanics}

\textbf{Planner-executor decomposition.} At the start of each layer $i$, the \textit{Planner} generates up to $n$ natural-language exploration questions $q_{i,1}, \cdots, q_{i,n}$, which may be directed at either the database $\mathcal{D}$ or the internet corpus $\mathcal{I}$. For the first layer, questions are generated from the preliminary report $r_0$ (Prompt~\ref{prompt:initial_questions}); for subsequent layers, they are generated from the current insight bank $B_{i-1}$ and thesis $t_{i-1}$ (Prompt~\ref{prompt:exploration_question_direct_gen}).

The \textit{Planner} only formulates higher-level exploratory questions, which are then translated to formal queries and executed by a separate \textit{Executor} agent. This abstraction allows for substitution of general text-to-SQL approaches. In this paper, we use a ReAct-style~\citep{yao2023react} agentic system based on \citet{liu-etal-2024-spinach} (see Appendix~\ref{appendix:executor-agent-details} for details). The \textit{Executor} is responsible for low-level query translation, and does so by dynamically exploring database contents to determine the most appropriate SQL response. This decomposition is necessary because, in complex databases, determining the correct query often requires agentic loops with feedback to handle schema-specific nuances~\citep{pourreza2023dinsqldecomposedincontextlearning, deng2025reforcetexttosqlagentselfrefinement, hua2026sqltrailmultiturnreinforcementlearning}. The \textit{Executor} returns a list of natural-language answers $a_{i,1}, \cdots, a_{i,n}$, each grounded in an executed SQL query $s_{i,j}$ that serves as the semantic parse of $q_{i,j}$.

\textbf{Query consistency module.} A key challenge when working with large-scale knowledge bases is that independent, parallel explorations can produce inconsistent queries. For instance, in the ACLED database, multiple columns encode the actors involved in an event, including \texttt{actor1}, \texttt{assoc\_actor\_1}, \texttt{actor2}, and \texttt{assoc\_actor\_2}. When analyzing specific actors, independent \textit{Executor} runs may handle these columns differently. For example, one run might filter using \texttt{`Protesters' in actor1 or `Protesters' in actor2}, while another includes associated actors via \texttt{`Protesters' in actor1 or `Protesters' in actor2 or `Protesters' in assoc\_actor\_1 or `Protesters' in assoc\_actor\_2}. Such discrepancies can result in entirely different aggregated statistics.

To address this, \system introduces a Query Consistency Module (Prompt~\ref{prompt:query_consistency_module}), which takes as input the database-directed question-query pairs $(q_{i,j}, s_{i,j})$ and produces consistency follow-up queries $q'_{i,j}$. The module also enforces global consistency by referencing question-query pairs stored in $B_{i-1}$, which are not modified but serve as context to guide standardization of the $q_{i,j}$. The follow-up queries $q'_{i,j}$ are then passed to the \textit{Executor}, together with $(q_{i,j}, s_{i,j})$ as context, to produce the final queries $s'_{i,j}$ and answers $a'_{i,j}$.

\textbf{Bottom-up inductive insight surfacing}: In EDA, the distinction between deductive reasoning (interrogating data based on a prior hypothesis) and inductive reasoning (exploring data openly to see what emerges) is well-established \citep{kirk2016data}. While the \textit{Planner} is primarily focused on deductive, top-down reasoning, it faces two practical limitations: it cannot investigate all queries of interest, and the tables returned by the \textit{Executor} often must be truncated due to length constraints. To address this, \system automatically computes summary statistics for each column in the returned table, embedding them directly into each answer $a'_{i,j}$. These include the distinct percentage (ratio of unique values to total rows), top-5 values, and for numerical columns, the minimum, maximum, median, and mean. For instance, consider the query shown in the bottom-left corner of Figure~\ref{fig:figure1}: the summary statistics inductively surface a maximum value significantly higher than the mean and median, which could prompt the \textit{Planner} to subsequently investigate the high-incident months in its next deductive query.

At the end of each layer, \system consolidates the new evidence surfaced by $a'_{i,j}$ with the existing insights in $B_{i-1}$, retaining up to a pre-configured maximum number of insights. Lower-quality insights already in $B_{i-1}$ may be displaced by stronger new findings (Prompt~\ref{prompt:insight_bank_filter}), producing the updated insight bank $B_i$.

\subsubsection{Thesis-Driven Exploration}

Relying solely on LLMs to explore topics of interest often leads to scattered insights without a central theme. In data storytelling, \citet{dykes2019effectivedatastorytelling} argues that ``it is equally important to engage your audience with a narrative---to tell a story with the numbers.'' We take this principle a step further: rather than reserving narrative for the final report, \system organizes its explorations around a guiding thesis from the outset. After the $p$-th layer, \system invokes a thesis generation module (Prompt~\ref{prompt:thesis_generation}), which takes the current insight bank $B_p$ as input and formulates an initial exploration thesis $t_p$. This thesis is then refined every $p$ layers thereafter (Prompt~\ref{prompt:thesis_refinement}), producing intermediate theses $t_{2p}, t_{3p}, \cdots$, steering subsequent exploration toward a coherent narrative.

\subsection{Final Report Generation}

\begin{figure}[t!]
  \centering
  \includegraphics[scale=0.52]{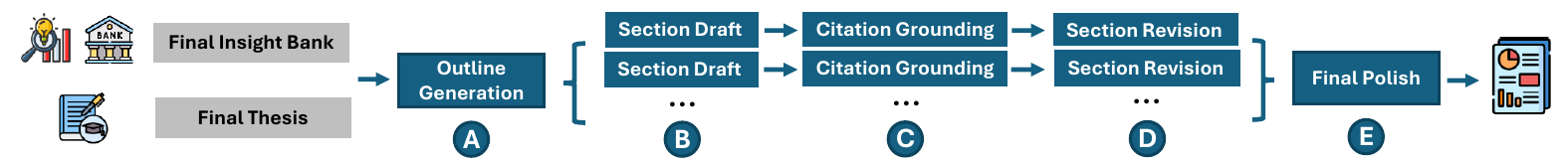}
  \caption{Overview of the Final Report Generation Module}
  \label{fig:final_report_gen}
\end{figure}

State-of-the-art LLMs, even with high reasoning effort, struggle to generate 
well-structured reports in a single pass \citep{shao-etal-2024-assisting, 
jiang-etal-2024-unknown, gu-etal-2025-rapid, gao-etal-2023-enabling}: they tend to produce claims 
ungrounded in citations, pad responses with long citation lists, and lack the 
coherent narrative structure needed to integrate evidence progressively toward a 
thesis.

To counter these tendencies, \system employs a staged editorial pipeline to produce 
the report $r$ from the final insight bank $B_m$ and thesis $t_m$. The pipeline 
proceeds in five stages, as illustrated in Figure~\ref{fig:final_report_gen}. First, 
the system generates a section-level outline in which each section $s$ specifies its 
narrative purpose, required evidence $B^s \subseteq B$, and key points (Stage A, 
Prompt~\ref{prompt:outline_gen}). Each section is then drafted independently, 
grounded in its curated evidence subset $B^s$ (Stage B, 
Prompt~\ref{prompt:section_draft}). Next, every drafted section undergoes sentence-level fact verification: the section is chunked at citation boundaries, and the LLM is asked to flag any claims not adequately supported by the cited source (Stage C, Prompt~\ref{prompt:citation_grounding}). This step is essential: without it, LLM-generated related work sections are known to contain large numbers of fabricated citations~\citep{linardon2025citation, bienz2026casemysteriouscitations}.
The collected  criticisms are then used to revise each section (Stage D, 
Prompt~\ref{prompt:section_revision}). Finally, the revised sections are aggregated 
and polished into the final report (Stage E, Prompt~\ref{prompt:final_polish}).

%% file: 4_experiments.tex
\section{Evaluation}

\subsection{Evaluation on existing benchmarks}

Existing benchmarks targeting database-driven exploration are limited~\citep{majumder2025discoverybench,DBLP:conf/iclr/SahuPRACDTZLVCP25}. Among these, we focus on InsightBench~\citep{DBLP:conf/iclr/SahuPRACDTZLVCP25}, a benchmark designed to evaluate business analytics agents on insight generation. For further discussion of dataset selection, see Appendix~\ref{appendix:other-benchmarks}. InsightBench comprises 100 benchmark instances, each consisting of a natural language goal, a database, and a set of labeled gold insights. Given an input goal (e.g., ``Find discrepancies and imbalances in the distribution of incidents across categories.'') and a database, the benchmark evaluates whether a system's generated insights align with the gold insights. Generated insights are compared against gold insights via LLM-as-a-judge: each gold insight's match score is defined as its highest match score across all predicted insights, yielding ``Insight-level'' scores. In addition, a system generates a summary from its insights, and the ``Summary-level'' score is derived by comparing this generated summary against a labeled reference summary.

The original paper~\citep{DBLP:conf/iclr/SahuPRACDTZLVCP25} uses Llama-3 as the judge; however, in our experiments, Llama-3 nearly always assigns a score of 7/10 copied verbatim from the evaluation prompt (Prompt \ref{prompt:insightbench-eval}), even when the insights clearly do not match. We therefore evaluate using alternative LLM, both open-source (Qwen-3-30B) and closed-source (GPT-4o), as judges to ensure meaningful comparisons and reproducibility.

We set the maximum number of layers to $m=5$, with $2$ queries at the first layer and $5$ queries at each subsequent layer in \system. For the summary-level score, we employ a simple prompt (Prompt~\ref{prompt:summary_prompt}) to summarize the predicted insights. We compare against the state-of-the-art system AgentPoirot~\citep{DBLP:conf/iclr/SahuPRACDTZLVCP25}, a ReAct-style agent that uses Python code to iteratively discover insights across 3 parallel runs.  As shown in Table~\ref{tab:insight_bench}, \system substantially outperforms AgentPoirot: by 14.8\% in insight-level recall and 5.9\% in summary-level recall under GPT-4o evaluation, and by 19.4\% and 7.2\% under Qwen3-30B evaluation, respectively.

\begin{table}[!t]
\centering
\resizebox{\columnwidth}{!}{%

\begin{tabular}{lcccc}
\toprule
\textbf{Method} 
& \multicolumn{2}{c}{\textbf{Insight-level}} 
& \multicolumn{2}{c}{\textbf{Summary-level}} \\
\cmidrule(lr){2-3} \cmidrule(lr){4-5}
& \textbf{GPT-4o judge} & \textbf{Qwen3-30B judge} 
& \textbf{GPT-4o judge} & \textbf{Qwen3-30B judge} \\
\midrule

\rowcolor{gray!20} \multicolumn{5}{c}{Previous SOTA} \\
AgentPoirot (gpt-5)  & 47.1 & 49.9 &46.6  & 51.5  \\

\midrule
\rowcolor{gray!20} \multicolumn{5}{c}{Our Approach} \\

DataSTORM (gpt-5)    & \textbf{61.9} & \textbf{69.3} & \textbf{52.5} &\textbf{ 58.7}  \\

\bottomrule
\end{tabular}
}
\caption{Insight- and summary-level recall (\%) on InsightBench~\citep{DBLP:conf/iclr/SahuPRACDTZLVCP25}, evaluated under two LLM judges. \system outperforms the state-of-the-art AgentPoirot consistently under both judges. Qwen3-30B refers to \texttt{Qwen/Qwen3-30B-A3B-Instruct-2507}.}
\label{tab:insight_bench}
\end{table}

\textbf{Error Analysis}: We conduct an error analysis by randomly sampling 20 gold-labeled insights with a maximum match score of 0.5. Among these, we find that 45\% stem from annotation errors in the dataset, including unsupported claims (e.g., references to ``printers'' absent from the data), unjustified conclusions (e.g., labeling cities such as Tokyo or London as high-retention without evidence), figure--label mismatches, and poor grounding in the input. This annotation quality issue has also been noted in prior work~\citep{zhu2025insightevalexpertcuratedbenchmarkassessing}. The remaining 55\% reflect genuine model failures to explore relevant analytical directions. Notably, 45\% of these genuine failures involve time-series trend analysis, which proves particularly challenging.

\subsection{Armed Conflict Location \& Event Data (ACLED) dataset}

Existing benchmarks do not capture the real complexity of deep research over both the internet and local databases in real-world settings. We therefore construct a new evaluation dataset built on top of the Armed Conflict Location \& Event Data (ACLED)~\citep{doi:10.1177/0022343310378914} database. This database contains manually labeled incidents of violent conflict and protest across the world since 2005, with rigorous labeling guidelines~\citep{acled_codebook_2024}. Conflict analysis naturally requires both internet sources and structured annotated databases, making ACLED a well-suited evaluation target.

\textbf{Dataset construction}: We use the event data published on the ACLED website~\footnote{\url{https://acleddata.com/conflict-data/download-data-files}, accessed 01/02/2026}. We use the publicly available portion of the dataset, which is subject to a one-year lag. The resulting dataset contains 1,630,187 rows spanning from 2020-01-01 to 2025-01-02 across 31 columns~\citep{acled_codebook_2024}. We then examine ACLED-expert-written analyses published in the latter half of 2024\footnote{\url{https://acleddata.com/global-analysis\#listing-20401}, accessed 01/02/2026}, collecting 20 articles covering events through the end of 2024 to make full use of the available data. Since article titles often reveal research conclusions, the authors manually reviewed each article and derived a neutral prompt suitable as input for deep research systems. See Appendix~\ref{appendix:acled-articles-list} for the full list of articles, links to the originals, and the corresponding neutral prompts.

\textbf{Baseline and Experiment Setup}: We compare against the strong proprietary system OpenAI Deep Research~\citep{openai2025deepresearch}, which has demonstrated state-of-the-art performance across a wide range of benchmarks~\citep{shao2025dr, du2025deepresearchbenchcomprehensivebenchmark}, under current API constriants. To ensure a fair comparison in which no system can access the ACLED article directly, we implement a custom web search server via Serper that blocks access to the ACLED domain. Each web server is further restricted to sites published prior to the corresponding ACLED article's publication date, preventing any leakage of future information. For database access, we experiment with two settings\rev{~(Appendix~\ref{appendix:acled-baseline-config})}:
\begin{enumerate}
    \item \textbf{OpenAI DR (CSV)}: We provide the system direct access to the raw CSV as an uploaded container file. Due to input size constraints, the CSV is chunked by region, which in practice simplifies the task. The model interacts with the data by writing Python code and observing the outputs.
    \item \textbf{OpenAI DR (MCP)}: We provide the system access to our database agent described in Section~\ref{sec:multi-agent-exploration-framework} via an MCP server~\citep{mcp_openai_docs_2025}, allowing the model to interact with the database through natural language queries and receive natural language responses.
\end{enumerate}

Both OpenAI Deep Research baselines use \texttt{o3-deep-research-2025-06-26}\footnote{The Deep Research API currently restricts usage to \texttt{o3-deep-research} and \texttt{o4-mini-deep-research} models.}. \system uses \texttt{gpt-5-2025-08-07} for multi-agent exploration and \texttt{gpt-5.1-2025-11-13} for final report drafting, fact-checking, and polishing.

\subsubsection{Automatic Evaluations}

Evaluating deep research systems poses significant challenges due to the open-ended nature of the task and the absence of a single ground truth~\citep{du2025deepresearchbenchcomprehensivebenchmark, patel2026deepscholarbenchlivebenchmarkautomated, futuresearch2025deepresearchbenchevaluating, xu2025researcherbenchevaluatingdeepai}. We use a wide variety of automated metrics, both from prior literature and newly developed by us, to evaluate the following dimensions:

\begin{enumerate}
\item \textit{Reference-induced criteria matching}: Given a reference article $a_r$ and a system-generated article $a_s$, we first prompt an LLM to derive a set of evaluation criteria from $a_r$ (example criteria is shown in Table~\ref{prompt:example_criteria}) using Prompt~\ref{prompt:criteria-matching-gen}. Each criterion is then used to grade $a_s$, deriving a score in $\{0, 0.25, 0.5, 0.75, 1.0\}$ via Prompt~\ref{prompt:criteria-matching-grade}, and scores are averaged across all criteria.
\item \textit{RACE evaluation framework}: Proposed by \citet{du2025deepresearchbenchcomprehensivebenchmark}, RACE is a reference-based, dynamically weighted evaluation framework that uses task-adaptive criteria and LLM-as-a-judge comparisons against expert reports to assess the quality of long-form research outputs: Comprehensiveness, Depth, Instruction Following, and Readability. Scores are defined relative to a reference report, where 50 denotes parity; scores above 50 indicate that the evaluated report is judged superior to the reference under task-specific weighted criteria.
\item \textit{Database use}: We evaluate the extent to which a system-generated article makes use of the underlying database. We first decompose each system-generated article $a_s$ into atomic claims using Prompt~\ref{prompt:atomic-breakdown}, then determine for each claim whether it is attributed to the ACLED database or to generic internet sources using Prompt~\ref{prompt:insight-attribution}.
\end{enumerate}

\begin{table}[!t]
\centering
\small
\begin{tabularx}{\linewidth}{l c *{5}{>{\centering\arraybackslash}X} c}
\toprule
 & \multicolumn{1}{c}{Ref-Induced}
 & \multicolumn{5}{c}{RACE Eval Framework~\citep{du2025deepresearchbenchcomprehensivebenchmark}}
  & \multicolumn{1}{c}{DB Use}\\
\cmidrule(lr){2-2} \cmidrule(lr){3-7} \cmidrule(lr){8-8}
\textbf{Approach} 
& \textbf{Avg Match}
& \textbf{Overall} 
& \textbf{Comp.} 
& \textbf{Depth} 
& \textbf{Inst.} 
& \textbf{Read.}
& \textbf{Ratio} \\
\midrule
OpenAI DR (MCP)          & 48.5\% & 46.1 & 45.6 & 42.8 & 49.3 & 49.3 & 30.4\%  \\
OpenAI DR (CSV)          & 51.2\% & 46.8 & 46.9 & 43.9 & 49.4 & 49.5 & 23.3\%\\
DataSTORM                & \textbf{61.8\%} & \textbf{52.6} & \textbf{52.5} & \textbf{52.2} & \textbf{53.5} & \textbf{52.5} & \textbf{66.4}\% \\
\bottomrule
\end{tabularx}
\caption{Comparison of OpenAI Deep Research and DataSTORM on the ACLED benchmark using reference-induced criteria matching and the RACE evaluation framework from \citet{du2025deepresearchbenchcomprehensivebenchmark}. All scores are evaluated using GPT-5. }
\label{tab:automatic-metrics}
\end{table}

As shown in Table~\ref{tab:automatic-metrics}, \system outperforms both OpenAI DR baselines across all metrics, achieving a 10.6\% improvement in reference-induced matching and a 5.8-point gain in overall RACE score, while incorporating 36\% more database content.

\rev{\subsubsection{Ablations and Analysis}
\label{sec:ablations}

In this section, we carry out additional ablation and manual analysis to understand the effect of each of \system's components. We ablate three modules, including (i) thesis generation, (ii) inductive bottom-up statistics layer, and (iii) query consistency module, and evaluate them with the paper's automatic metrics on a sample of 10 topics (ID: 1, 2, 4, 8, 9, 10, 11, 12, 17, 18).

\textbf{Both the inductive statistics layer and thesis generation contribute meaningfully to \system's performance.} As shown in Table~\ref{tab:ablation}, removing thesis generation lowers reference-induced criteria matching by 13.5 points, and removing the bottom-up inductive statistics layer lowers it by 5.4 points, each accompanied by a decline across the RACE dimensions. Additional experiments analyzing how the thesis refinement module mitigates confirmation bias are presented in Appendix \ref{appendix:thesis-refinement-analysis}.

\textbf{Query consistency layer contributes meaningfully to \system’s performance.} The query consistency module standardizes SQL filters across a report so that its numerical claims are mutually comparable. We evaluate this behavior with a dedicated query-consistency metric. The metric asks: when multiple SQL queries in the same report refer to the same entity, geography, time period, event category, or counting target, do they operationalize that concept in the same way? Refer to Appendix ~\ref{appendix:query-consistency-metric} for details on metric construction, scoring, and representative flagged inconsistencies. The result on 10 topics are shown in Table~\ref{tab:query-consistency-results}. This result shows that enabling query consistency raises the overall score from 0.406 to 0.549, and full DataSTORM is the more consistent system on five of the six dimensions.

Additional analysis on \system's numerical faithfulness is presented in Appendix \ref{appendix:claim-correctness}.

}

\begin{table}[t]
\centering
\small
\setlength{\tabcolsep}{5pt}
\begin{tabular}{lccc}
\toprule
Rubric & OpenAI DR (CSV) & \system & Win\% (lose\%) \\
\midrule
Public Impact & 3.58 & \textbf{3.60} & 25.0\% (20.0\%) \\
Public Interest & 3.30 & \textbf{3.50} & 32.5\% (20.0\%) \\
Institutional Significance & 3.45 & \textbf{3.58} & 30.0\% (17.5\%) \\
Conflict and Stakes & 3.43 & \textbf{3.83} & 45.0\% (22.5\%) \\
Originality & 2.21 & \textbf{3.05$^{\dagger}$} & 61.5\% (10.3\%) \\
Human Consequences & 2.75 & \textbf{2.88} & 37.5\% (35.0\%) \\
Follow-up Potential & 3.38 & 3.38 & 27.5\% (27.5\%) \\
\bottomrule
\end{tabular}
\caption{Human ratings of newsworthiness dimensions for reports generated by \system and OpenAI DR (CSV) ($N=40$ reviewer-topic evaluations; $N=39$ for \textit{Originality} because one evaluator marked one topic as \textit{N/A}). Ratings use a 1--5 scale. Evaluators could assign \textit{N/A} when an article did not provide sufficient basis for judgment; such cases were excluded from the corresponding statistics. We report mean ratings and the pairwise win rate of \system. $\dagger$ indicates a significant difference ($p<0.05$) between \system and OpenAI DR (CSV) in a paired t-test.}
\label{tab:rubric_openai_vs_system}
\end{table}

\subsubsection{Human Evaluation}

We further conduct an IRB-approved human evaluation on the ACLED dataset. Because ACLED centers on conflict analysis and report framing closely aligned with investigative journalism, we recruited 10 evaluators with investigative-reporting experience online. On average, participants had 13.3 years of journalism experience. We compare \system against the strongest automatic baseline, OpenAI DR (CSV). Additional details of the study protocol, evaluator recruitment, and interface are provided in Appendix~\ref{appendix:human-evaluation-details}.

\paragraph{Evaluation Setup}
For each of the 20 topics in the dataset, we assigned two participants. Participants were first shown a neutral seed topic derived from the corresponding ACLED reference article. This seed topic provides only high-level context and scope, without revealing the article's framing or conclusion. Participants then read the expert-written ACLED article as a reference for how a human expert framed the topic and selected a thesis. Afterward, they evaluated the outputs of \system and OpenAI DR (CSV). To minimize presentation bias, we anonymized both system outputs and standardized their presentation style. Evaluators were not informed which system produced which article. To counterbalance order effects, one participant viewed \system before the baseline, while the other viewed the systems in the reverse order.

For each system output, participants rated seven rubric items on a 5-point Likert scale. The rubric was adapted from classic newsworthiness criteria \citep{galtung1965structure} and covers public impact, public interest, institutional significance, conflict and stakes, originality, human consequences, and follow-up potential. Participants also reported pairwise preferences over thesis strength and reporting angle for all three report pairs. Optional free-form feedback was collected.

\paragraph{Human Evaluation Results} As shown in Table \ref{tab:rubric_openai_vs_system}, \system outperforms the OpenAI DR (CSV) baseline across 6 dimensions of newsworthiness, driven primarily by its capacity for thesis-driven research rather than surface-level summarization. This paradigm shift is most evident in the \textit{Originality} dimension where \system achieves a statistically significant improvement. Qualitative coding of free-form feedback reveals that evaluators praised \system for establishing a clear, distinctive analytical angle in 32.5\% of comments. In contrast, the baseline was criticized in 40.0\% of comments for relying on weak theses and passive literature summaries. Unlike the baseline, \system actively mines raw knowledge sources to propose, defend, and iterate upon concrete hypotheses.

\begin{figure*}[t]
\centering
\includegraphics[width=.7\textwidth]{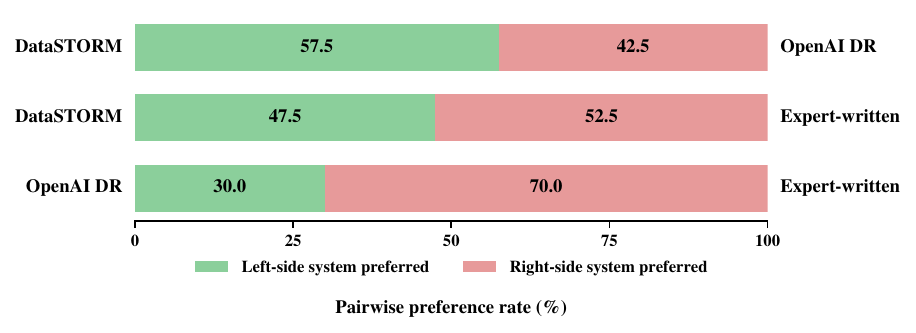}
\caption{Pairwise preference rates in human evaluation. For each comparison, the bar shows the percentage of evaluations preferring the left-side system versus the right-side system.}
\label{fig:human_pref}
\vspace{-10pt}
\end{figure*}

\system has better analytical framing that grounds underlying data in concrete human stakes. Evaluators found that \system successfully isolated consequential events to organize reports around coherent arguments, which is reflected in its higher scores for \textit{Conflict and Stakes}. Qualitative feedback explicitly praised \system for making human consequences concrete, whereas the baseline was frequently criticized (25\% of comments) for lacking human voice or failing to articulate the immediate relevance of the events. This pattern is also reflected in the pairwise preferences in Figure~\ref{fig:human_pref}. evaluators preferred \system over OpenAI DR in 57.5\% of comparisons. \system also remained close to expert-written reports, receiving a 47.5\% preference rate against them, whereas OpenAI DR was preferred over expert-written reports in only 30.0\% of comparisons.

Evaluator feedback highlights a critical trade-off regarding presentation style and reader orientation. While acknowledging \system's strong underlying arguments, evaluators noted limitations in its readability, accessibility, and communication of downstream reporting directions in 27.5\% of comments. Conversely, the baseline was frequently praised (20\% of comments) for providing helpful background context and smooth reader onboarding. This underscores that coupling \system's thesis-driven research capabilities with the clean, accessible, and context-rich reporting style remains a critical direction for future work.

%% file: 6_conclusion.tex
\section{Conclusion}
This paper introduces \system, a deep research system capable of autonomously exploring large-scale databases and internet sources. Grounded in the practices of Exploratory Data Analysis and Data Storytelling, \system combines planner-executor decomposition, inductive and deductive reasoning, and thesis-driven exploration to produce coherent, evidence-backed narratives. We evaluate \system against state-of-the-art baselines,  AgentPoirot and OpenAI Deep Research, on existing benchmarks and a novel real-world dataset built on ACLED conflict data, demonstrating that \system substantially outperforms both. We hope \system serves as a foundation for future work in autonomous, data-driven discovery.

%% file: 8_limitations_ethical_considerations.tex
\section*{Acknowledgment}

The authors would like to acknowledge Katayoun Kishi, Trey Billing, and Clionadh Raleigh of ACLED for their insightful discussions and contributions; Kaveh Waddell and Eryn Davis for their feedback on the user study; and Harshit Joshi, Jiuding Sun, Tamara Czinczoll, Yanzhe Zhang, Yijia Shao, Abhinav Chinta, Leah Harrison, Liana Patel, and other members of the Stanford Open Virtual Assistant Lab (OVAL) and NLP group for their feedback and support.

This work is supported in part by the Verdant Foundation, the Hasso Plattner Institute, Itaú Unibanco, BMO Financial Group, the Stanford Human-Centered Artificial Intelligence (HAI) Institute, and the Brown Institute for Media Innovation. We acknowledge the National Artificial Intelligence Research Resource (NAIRR) Pilot and Microsoft Azure for contributing to the results in this work. Shicheng Liu is supported by an IBM PhD Fellowship and gratefully acknowledges his mentor, Marquita Ellis, for her insightful discussions and guidance related to this work.

\section*{Reproducibility Statement}

We open-source our code, prompts, data, and experimental artifacts at \url{https://github.com/stanford-oval/DataSTORM}.

%% file: 99_appendix.tex
\newpage
\appendix
\section*{Appendix}
\startcontents[sections]
\printcontents[sections]{l}{1}{\setcounter{tocdepth}{2}}
\newpage

\section{Limitations and Future Work}
Data cleaning, or referred to in Exploratory Data Analysis literature as data ``wrangling'', is beyond the scope of this paper. DataSTORM expects a cleaned dataset as input and we leave data cleaning with LLMs as future research~\citep{zhang2025datascibenchllmagentbenchmark}.  Additionally, the Planner inherits the hypothesis-generation biases of its underlying LLM, including selection and framing biases, and developing more diverse exploration strategies is a natural direction for future research.  Finally, improving time-series reasoning, which accounts for 45\% of genuine model failures, and enhancing report readability are open challenges we leave to future work.

\section{Details on the Executor agent}
\label{appendix:executor-agent-details}

We create a ReAct-style database agent with the following available actions:

\begin{enumerate}
    \item \texttt{get\_tables()}: Retrieves all tables along with a brief description of each.
    \item \texttt{retrieve\_tables\_details([table\_1, table\_2, $\cdots$])}: Retrieves detailed information about the specified tables \texttt{table\_1, table\_2, $\cdots$}.
    \item \texttt{execute\_sql(sql)}: Executes a SQL query and returns the results.
    \item \texttt{execute\_python\_from\_sql(sql, python\_code)}: Executes Python code over the results of a SQL query.
    \item \texttt{stop()}: Designates the most recently executed SQL query as the final answer and terminates the process.
\end{enumerate}

At each turn, the agent inspects the current observations, selects one of the above actions with appropriate arguments, and the system executes it and appends the resulting observation to the context for the next turn. The main ReAct prompt is given in Prompt~\ref{prompt:executor-main}. The agent runs until it invokes \texttt{stop()}, or until 15 turns have elapsed, whichever comes first.

\section{Human evaluation details}
\label{appendix:human-evaluation-details}
\paragraph{Participants}
We recruited 10 evaluators with journalism experience through online outreach. On average, participants reported 13.3 years of experience in journalism. In terms of experience distribution, 3 participants reported at most 5 years of experience, 4 reported between 6 and 15 years, and 3 reported more than 15 years. Participants also voluntarily reported their highest education level: 4 held a bachelor's degree, 5 held a master's degree, and 1 held a PhD.

\begin{figure}[t]
    \centering
    \includegraphics[width=\linewidth]{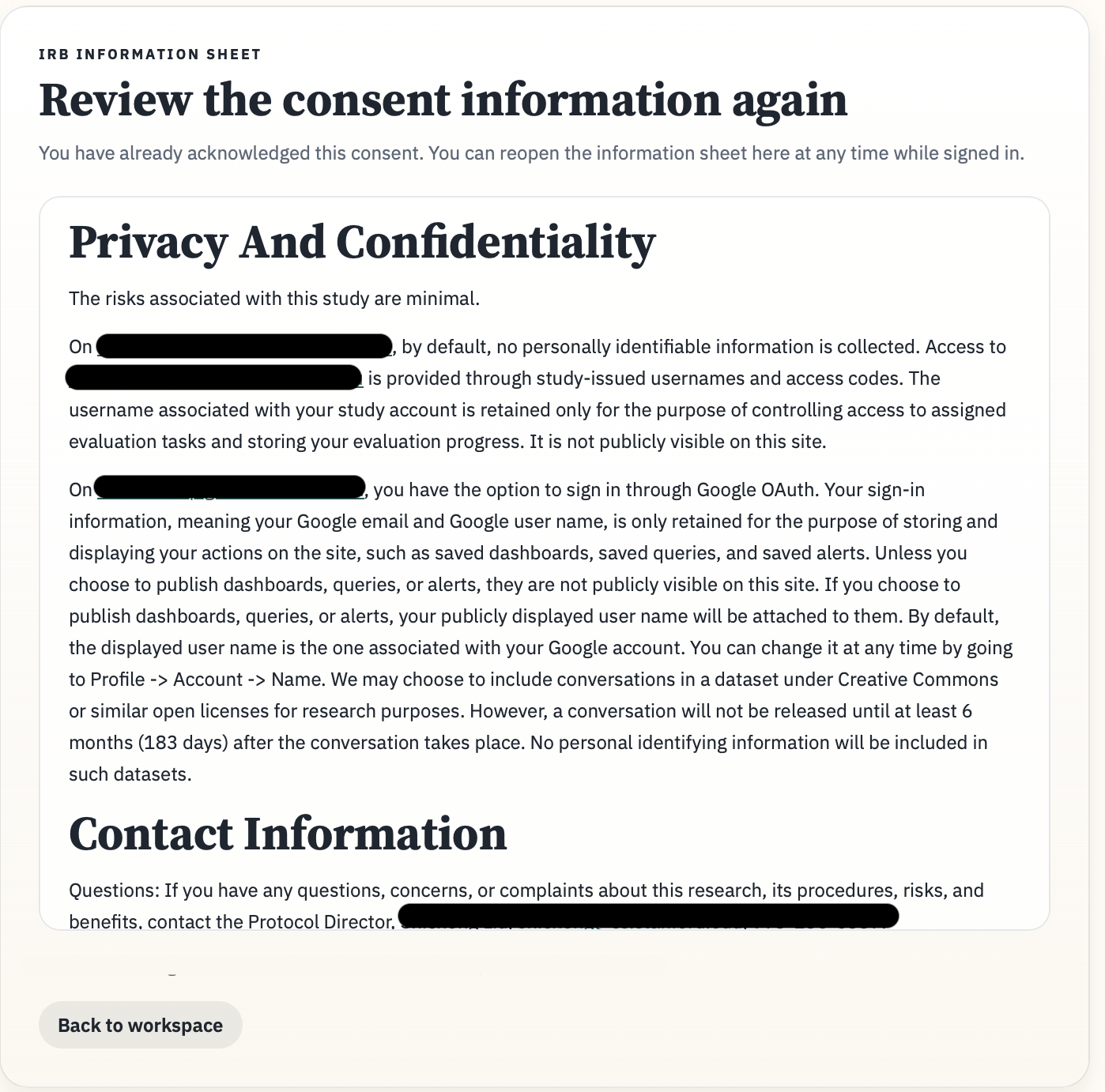}
    \caption{Consent page shown to participants before the human evaluation. Identifying information has been redacted for anonymous review.}
    \label{fig:human-eval-consent}
\end{figure}

\paragraph{Ethics and data handling}
All participants provided informed consent before beginning the study. The consent form described the purpose of the study, the voluntary nature of participation, the expected time commitment, and our data-handling practices. We did not store personally identifiable information as part of the study data. The study protocol and consent materials were reviewed under our IRB process. Figure~\ref{fig:human-eval-consent} shows the consent page presented to participants.

\begin{figure}[t]
    \centering
    \includegraphics[width=\linewidth]{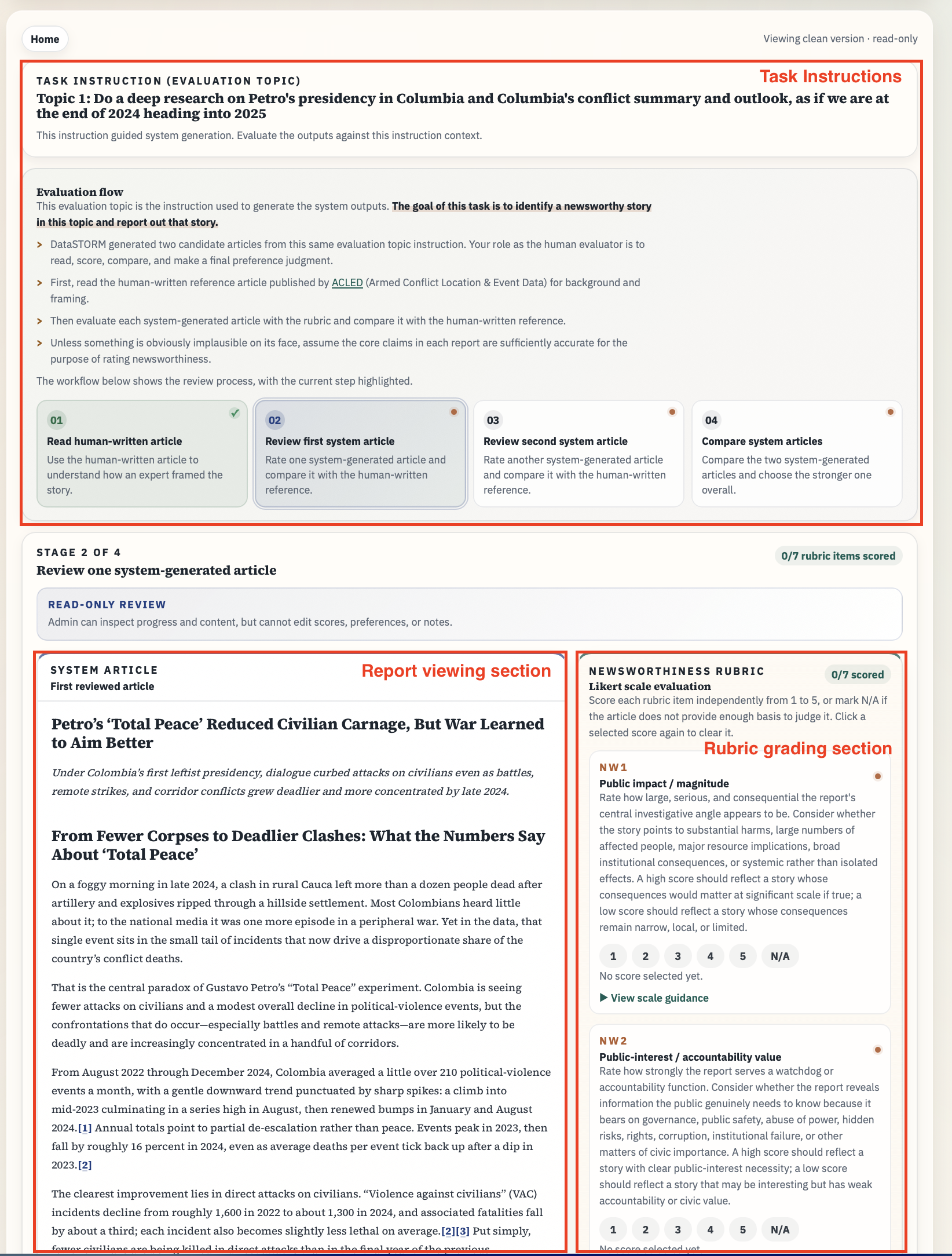}
    \caption{Screenshot of the custom web interface used for human evaluation. Participants reviewed the reference article and system outputs, assigned rubric-based newsworthiness ratings, and provided pairwise comparisons and optional free-form comments.}
    \label{fig:human-eval-interface}
\end{figure}

\paragraph{Evaluation interface and procedure}
Feedback was collected through a custom web application developed for this study. For each topic, participants were first shown task instructions, a neutral seed topic, and the corresponding expert-written ACLED article. The ACLED article served as a reference for how a human expert framed the topic and selected a reporting thesis.

Participants then evaluated one system-generated report at a time. After reading a report, they assigned 1--5 Likert ratings on seven newsworthiness rubrics (Listing~\ref{lst:newsworthiness-rubric}), provided a pairwise comparison against the human expert-written article with respect to reporting angle, and could optionally leave free-form comments. They then repeated the same process for the second system-generated report. After reviewing both system outputs, participants provided a final pairwise comparison between the two systems and could optionally explain their rationale in free-form text. Figure~\ref{fig:human-eval-interface} shows the evaluation interface.

\rev{\paragraph{Treatment of disagreement} The rubrics target subjective dimensions such as newsworthiness, thesis strength, and reporting angle, for which there is no single correct label. We therefore treat each evaluator's judgment as an independent assessment and report aggregate ratings and pairwise preference rates across evaluator-topic instances, rather than adjudicating disagreements into one label per topic.}

\section{Details on other existing datasets considered}
\label{appendix:other-benchmarks}

We also considered another benchmark referenced in the literature, DiscoveryBench~\citep{majumder2025discoverybench}. In preliminary experiments, we found that the dataset does not fully release the data required for evaluation. Specifically, only the ``real/train'' subset is available, while the ``real/test'' portion lacks usable evaluation signals. Furthermore, the gold labels in ``real/train'' do not include code or intermediate steps explaining how the labels were derived.

Upon manual inspection of 10 cases, we were unable to verify the computations underlying the gold labels, and observed that many appear to contain erroneous annotations.

\rev{

\section{Additional ablations and analysis}

\begin{table}[t]
\centering
\color{revcolor}
\small

\begin{subtable}{\linewidth}
\centering
\setlength{\tabcolsep}{4pt}
\begin{tabularx}{\linewidth}{l c *{5}{>{\centering\arraybackslash}X} c}
\toprule
 & \multicolumn{1}{c}{Ref-Induced}
 & \multicolumn{5}{c}{RACE Eval Framework~\citep{du2025deepresearchbenchcomprehensivebenchmark}}
 & \multicolumn{1}{c}{DB Use}\\
\cmidrule(lr){2-2}
\cmidrule(lr){3-7}
\cmidrule(lr){8-8}
\textbf{Approach}
& \textbf{Avg Match}
& \textbf{Overall}
& \textbf{Comp.}
& \textbf{Depth}
& \textbf{Inst.}
& \textbf{Read.}
& \textbf{Ratio} \\
\midrule
DataSTORM
& \textbf{62.3\%}
& \textbf{51.0}
& \textbf{51.2}
& \textbf{50.2}
& \textbf{51.8}
& 51.9
& 67.8\% \\

\quad w/o thesis generation
& 48.8\%
& 48.6
& 47.4
& 48.1
& 49.9
& 50.8
& \textbf{71.9}\% \\

\quad w/o inductive surfacing
& 56.9\%
& 50.1
& 50.1
& 48.7
& 51.4
& \textbf{52.1}
& 70.9\% \\
\bottomrule
\end{tabularx}
\caption{Leave-one-out ablations of \system.}
\label{tab:ablation}
\end{subtable}

\vspace{0.8em}

\begin{subtable}{\linewidth}
\centering
\setlength{\tabcolsep}{4pt}
\begin{tabularx}{\linewidth}{l c *{6}{>{\centering\arraybackslash}X}}
\toprule
\textbf{Approach}
& \textbf{Overall}
& \textbf{Pop.}
& \textbf{Event}
& \textbf{Actor}
& \textbf{Count}
& \textbf{Geo.}
& \textbf{Temp.} \\
\midrule
DataSTORM
& \textbf{0.549}
& \textbf{0.68}
& \textbf{0.53}
& \textbf{0.25}
& \textbf{0.50}
& \textbf{0.50}
& 0.83 \\

\quad w/o query consistency
& 0.406
& 0.25
& 0.36
& 0.10
& 0.45
& 0.45
& 0.83 \\
\bottomrule
\end{tabularx}
\caption{Ablation of query consistency module. Columns abbreviate the six dimensions of Table~\ref{tab:query-consistency-dimensions}.}
\label{tab:query-consistency-results}
\end{subtable}

\caption{Ablation results for DataSTORM on the 10 ACLED topics. All scores judged by GPT-5.}
\label{tab:combined-ablations}
\end{table}

\begin{table}[t]
\centering
\color{revcolor}
\small
\begin{tabular}{lccc}
\toprule
\textbf{Axis} & \textbf{Correct} & \textbf{Unsure} & \textbf{Incorrect} \\
\midrule
Faithfulness to cited result table & 86\% & 4\% & 10\% \\
Retrieval correctness of SQL query & 94\% & 6\% & 0\% \\
\bottomrule
\end{tabular}
\caption{Manual verification of 50 numerical, SQL-backed claims sampled from 18 \system reports on the ACLED benchmark.}
\label{tab:claim-correctness}
\end{table}

\subsection{Query consistency metric}
\label{appendix:query-consistency-metric}

\paragraph{Metric construction} For each report, the evaluator extracts every SQL-backed piece of evidence and represents it as a triple $(q_i, s_i, a_i)$, where $q_i$ is the natural-language question, $s_i$ is the SQL query, and $a_i$ is a short preview of the returned table. Identical SQL strings are deduplicated. All of a report's triples are then passed to GPT-5, which performs two steps within each of the six dimensions of Table~\ref{tab:query-consistency-dimensions}. In the \textit{grouping} step, it identifies concept groups, that is, sets of two or more queries that target the same underlying concept (e.g., the same actor, place, time period, or event category) within that dimension; purely structural predicates such as join conditions are ignored. In the \textit{consistency judgment} step, it decides for each group whether the queries operationalize that shared concept with semantically equivalent filters, and records the conflicting predicates for any group it finds inconsistent.

\paragraph{Scoring} Each dimension's score is the fraction of its multi-query concept groups whose filters are semantically consistent; a dimension with no comparable multi-query groups scores $1.0$, since no inconsistency is possible. A report's overall score is the mean of its six dimension scores. Table~\ref{tab:query-consistency-results} reports per-dimension scores over 10 topics.

\begin{table}[ht]
\centering
\color{revcolor}
\small
\begin{tabularx}{\linewidth}{lXX}
\toprule
\textbf{Dimension} & \textbf{What it checks} & \textbf{Example inconsistency} \\
\midrule
Actor / entity matching & Whether the same actor or group is matched the same way & One query searches only \texttt{actor1} and \texttt{actor2} for ``Fano,'' while another also searches \texttt{assoc\_actor\_*} and \texttt{notes} \\
\addlinespace
Geography & Whether the same place is normalized the same way & One query uses \texttt{admin1 = 'Amhara'}, while another uses \texttt{country = 'Ethiopia' AND admin1 = 'Amhara'} \\
\addlinespace
Temporal & Whether the same time period uses the same date window & One query uses \texttt{event\_date >= '2023-04-01'}, while another uses \texttt{BETWEEN '2023-04-01' AND '2024-12-31'} \\
\addlinespace
Event taxonomy & Whether the same event concept is defined consistently & ``Political violence'' is defined using \texttt{disorder\_type} in one query but \texttt{event\_type} in another \\
\addlinespace
Counting / aggregation & Whether the same quantity is counted the same way & One query reports \texttt{COUNT(*)} (number of events) while another reports \texttt{SUM(fatalities)} (death toll) for the same magnitude claim \\
\addlinespace
Population scope & Whether the same base population is filtered the same way & One query includes \texttt{country = 'Sudan'}, while another omits it for the same regional claim \\
\bottomrule
\end{tabularx}
\caption{The six ACLED-relevant dimensions of the query consistency metric, and the kind of inconsistency each one catches.}
\label{tab:query-consistency-dimensions}
\end{table}

\paragraph{Representative flagged inconsistencies} The following are inconsistencies the metric flags when the query consistency module is removed:
\begin{itemize}[leftmargin=*,itemsep=0pt]
\item \textit{Population scope}: one query restricts \texttt{country = 'Sudan'} while another omits it for the same regional claim, so the two counts are drawn from different base sets.
\item \textit{Event taxonomy}: ``political violence'' is defined as \texttt{disorder\_type IN ('Political violence', ...)} in one query but as \texttt{event\_type IN ('Battles', 'Explosions/Remote violence', ...)} in another, placing different universes of events behind the same term.
\item \textit{Temporal}: one query uses \texttt{event\_date >= '2023-04-01'} with no upper bound while others use \texttt{BETWEEN '2023-04-01' AND '2024-12-31'}, silently pulling data past the report's window.
\item \textit{Actor matching}: ``Fano'' is matched on \texttt{actor1}/\texttt{actor2} only in one query but also across \texttt{assoc\_actor\_*} and \texttt{notes} in another, changing which events count.
\end{itemize}

\subsection{Thesis refinement analysis}
\label{appendix:thesis-refinement-analysis}

The thesis refinement module keeps the working thesis revisable as evidence accumulates, choosing among the \textit{sharpen}, \textit{pivot}, and \textit{confirm} actions (Prompt~\ref{prompt:thesis_refinement}). Across the ACLED runs, the distribution of thesis refinement events was:

\begin{enumerate}
    \item Sharpen: 34
    \item Pivot: 3
    \item Confirm: 1
    \item Confirm-and-sharpen: 2
\end{enumerate}

These results suggest that the system does not simply confirm the initial thesis at every refinement step. Most refinements sharpened the thesis by adding specificity or constraints, while a smaller number explicitly pivoted the thesis in response to new evidence.

To further examine whether the final reports merely reproduced the initial thesis, we also conducted an LLM-as-a-judge analysis of the evidence included after the initial thesis was generated. For each evidence item in the final report, we asked the judge model to classify its relationship to the initial thesis as one of the following:

\begin{itemize}[leftmargin=*,itemsep=0pt]
\item \textit{Supports}: strengthens the thesis as written.
\item \textit{Qualifies}: adds an important caveat, boundary condition, exception, or scope limitation while leaving the main thesis partly intact.
\item \textit{Refutes}: directly contradicts a central claim or empirical premise.
\item \textit{Mixed}: contains both supporting and qualifying or refuting content.
\end{itemize}

Among 342 relevant evidence items, 267 supported the initial thesis, 59 qualified it, 3 refuted it, and 13 were mixed. Thus, 75 evidence items, or 21.9\%, introduced qualifications, contradictions, or mixed evidence relative to the initial thesis. This indicates that a substantial fraction of the evidence incorporated after thesis generation was not purely confirmatory, suggesting that the refinement process helps mitigate confirmation bias and premature lock-in.

\subsection{Correctness of numerical claims}
\label{appendix:claim-correctness}

\textbf{\system's numerical claims are grounded in the database evidence they cite.} We randomly sample 50 numerical, SQL-backed claims from 18 reports. For each claim, an author inspects the cited SQL query, the returned table, and the supporting rows, and labels it along two axes. \textit{Faithfulness} asks whether the number correctly reflects the cited result table, and \textit{retrieval correctness} asks whether the SQL actually answers the question. Each axis admits the labels correct, incorrect, and unsure.  As shown in Table~\ref{tab:claim-correctness}, 86\% are faithful, 10\% unfaithful, and 4\% unsure; on retrieval correctness, 94\% use a correctly retrieving query, 6\% are unsure, and none are substantively wrong. The high faithfulness and retrieval-correctness rates show that \system’s numerical claims are generally well supported by their cited SQL evidence. Importantly, we did not find any cases where \system’s SQL retrieved substantively wrong evidence for a numerical claim. Overall, this manual analysis provides direct evidence that \system’s reports are generally well grounded in the underlying database. It also highlights an important future work opportunity to further strengthen numerical faithfulness, building on the fact-checking module already incorporated in the \system system.
\section{ACLED baseline configuration}
\label{appendix:acled-baseline-config}

The reference ACLED articles postdate the neutral prompts derived from them, so restricting web search to sites published before each article's publication date prevents leakage of future information and ensures no system can reach the target article during evaluation. The two database-access settings serve complementary roles. The MCP setting gives OpenAI Deep Research the same database agent interface that \system uses, isolating the effect of the surrounding research framework, while the CSV setting gives it direct file access to the raw data and therefore does not depend on our interface at all. The gap in database use is not an artifact of the MCP interface. With direct CSV access, OpenAI DR attributes 23.3\% of its claims to the database, compared with 66.4\% for \system.
}

\section{ACLED Benchmark article list}
\label{appendix:acled-articles-list}

\textbf{ID}: 1
\begin{itemize}
\item \textbf{Original Article Name (with link)}: \href{https://acleddata.com/report/peace-talks-narino-may-be-litmus-test-petros-bid-end-colombias-conflict}{Peace talks in Nariño may be a litmus test for Petro’s bid to end Colombia’s conflict}
\item \textbf{Date Publish}: 12 December 2024
\item \textbf{Article Type}: Conflict Watchlist 2025; Report
\item \textbf{Neutral Deep Research Prompt}: ``Do a deep research on Petro's presidency in Columbia and Columbia's conflict summary and outlook, as if we are at the end of 2024 heading into 2025''
\end{itemize}

\textbf{ID}: 2
\begin{itemize}
\item \textbf{Original Article Name (with link)}: \href{https://acleddata.com/report/total-peace-paradox-colombia-petros-policy-reduced-violence-armed-groups-grew-stronger}{``Total Peace'' paradox in Colombia: Petro's policy reduced violence, but armed groups grew stronger}
\item \textbf{Date Publish}: 28 November 2024
\item \textbf{Article Type}: Report
\item \textbf{Neutral Deep Research Prompt}: ``Analyze Petro’s total peace policy in Colombia and its impact on violence v. armed groups.''
\end{itemize}

\textbf{ID}: 3
\begin{itemize}
\item \textbf{Original Article Name (with link)}: \href{https://acleddata.com/report/conflict-intensifies-and-instability-spreads-beyond-burkina-faso-mali-and-niger}{Conflict intensifies and instability spreads beyond Burkina Faso, Mali, and Niger
}
\item \textbf{Date Publish}: 12 December 2024
\item \textbf{Article Type}: Conflict Watchlist 2025; Report
\item \textbf{Neutral Deep Research Prompt}: ``Analyze levels of violence in Burkina Faso, Mali, and Niger, especially related to JNIM and IS Sahel operations.''
\end{itemize}

\textbf{ID}: 4
\begin{itemize}
\item \textbf{Original Article Name (with link)}: \href{https://acleddata.com/report/russias-protracted-war-ukraine-may-be-reaching-turning-point}{Russia’s protracted war on Ukraine may be reaching a turning point}
\item \textbf{Date Publish}: 12 December 2024
\item \textbf{Article Type}: Conflict Watchlist 2025; Report
\item \textbf{Neutral Deep Research Prompt}: ``Status of the Russian-Ukrainian conflict, as if we are at the end of 2024 heading into 2025.''
\end{itemize}

\textbf{ID}: 5
\begin{itemize}
\item \textbf{Original Article Name (with link)}: \href{https://acleddata.com/report/al-shabaab-targets-civilians-somalia-retaliation-installing-cctv-cameras-november-2024}{Al-Shabaab targets civilians in Somalia in retaliation for installing CCTV cameras - November 2024}
\item \textbf{Date Publish}: 29 November 2024
\item \textbf{Article Type}: Report
\item \textbf{Neutral Deep Research Prompt}: ``How is Al-Shabaab targeting civilians in Somalia following recent events?''
\end{itemize}

\textbf{ID}: 6
\begin{itemize}
\item \textbf{Original Article Name (with link)}: \href{https://acleddata.com/update/two-years-after-pretoria-agreement-unrest-still-looms-tigray-october-2024}{Two years after the Pretoria agreement, unrest still looms in Tigray - October 2024}
\item \textbf{Date Publish}: 8 November 2024
\item \textbf{Article Type}: Update
\item \textbf{Neutral Deep Research Prompt}: ``What is the status of conflicts in Tigray after the Pretoria agreement? Focus on October 2024.''
\end{itemize}

\textbf{ID}: 7
\begin{itemize}
\item \textbf{Original Article Name (with link)}: \href{https://acleddata.com/report/georgia-existential-election}{Georgia: An “existential” election}
\item \textbf{Date Publish}: 21 October 2024
\item \textbf{Article Type}: Election Watch; Report
\item \textbf{Neutral Deep Research Prompt}: ``Provide an analysis of Georgia's upcoming election, civil unrest, and political developments as of October 2024.''
\end{itemize}

\textbf{ID}: 8
\begin{itemize}
\item \textbf{Original Article Name (with link)}: \href{https://acleddata.com/report/artillery-shelling-and-airstrikes-surge-sudan-september-2024}{Artillery shelling and airstrikes surge in Sudan - September 2024}
\item \textbf{Date Publish}: 16 September 2024
\item \textbf{Article Type}: Report
\item \textbf{Neutral Deep Research Prompt}: ``Give me a summary of artillery shelling and airstrikes political violence, and conflict situations during August - September 2024 in Sudan.''
\end{itemize}

\textbf{ID}: 9
\begin{itemize}
\item \textbf{Original Article Name (with link)}: \href{https://acleddata.com/update/amhara-over-7-million-people-are-exposed-political-violence-august-2024}{In Amhara, over 7 million people are exposed to political violence - August 2024}
\item \textbf{Date Publish}: 13 September 2024
\item \textbf{Article Type}: Update
\item \textbf{Neutral Deep Research Prompt}: ``Status of political violence in Amhara since Fano insurgency began in April 2023.''
\end{itemize}

\textbf{ID}: 10
\begin{itemize}
\item \textbf{Original Article Name (with link)}: \href{https://acleddata.com/report/militants-thrive-amid-political-instability-pakistan}{Militants thrive amid political instability in Pakistan}
\item \textbf{Date Publish}: 12 December 2024
\item \textbf{Article Type}: Conflict Watchlist 2025; Report
\item \textbf{Neutral Deep Research Prompt}: ``What are the status of militants in Pakistan over the last year and what would be the outlook for conflicts involving militants in Pakistan going into 2025?''
\end{itemize}

\textbf{ID}: 11
\begin{itemize}
\item \textbf{Original Article Name (with link)}: \href{https://acleddata.com/report/between-cooperation-and-competition-struggle-resistance-groups-myanmar}{Between cooperation and competition: The struggle of resistance groups in Myanmar}
\item \textbf{Date Publish}: 26 November 2024
\item \textbf{Article Type}: Report
\item \textbf{Neutral Deep Research Prompt}: ``Research state of resistance groups in Myanmar post 2021 coup: What is the landscape of these groups?''
\end{itemize}

\textbf{ID}: 12
\begin{itemize}
\item \textbf{Original Article Name (with link)}: \href{https://acleddata.com/report/kenya-battles-threats-communal-militias-and-al-shabaab-november-2024}{Kenya battles threats from communal militias and al-Shabaab - November 2024}
\item \textbf{Date Publish}: 25 November 2024
\item \textbf{Article Type}: Report
\item \textbf{Neutral Deep Research Prompt}: ``What are the threats in Kenya with respect to communal militias and al-Shabaab. Focus on November 2024.''
\end{itemize}

\textbf{ID}: 13
\begin{itemize}
\item \textbf{Original Article Name (with link)}: \href{https://acleddata.com/update/cabo-ligado-update-11-24-november-2024}{Cabo Ligado Update: 11 - 24 November 2024}
\item \textbf{Date Publish}: 28 November 2024
\item \textbf{Article Type}: Mozambique Conflict Monitor - Cabo Ligado; Update
\item \textbf{Neutral Deep Research Prompt}: ``Research clashes between Islamic State Mozambique and state forces in Cabo Delgado. Focus on November 2024.''
\end{itemize}

\textbf{ID}: 14
\begin{itemize}
\item \textbf{Original Article Name (with link)}: \href{https://acleddata.com/update/year-after-snnprs-dissolution-violence-returns-historically-troubled-areas-september-2024}{A year after SNNPR's dissolution, violence returns to historically troubled areas - September 2024}
\item \textbf{Date Publish}: 17 October 2024
\item \textbf{Article Type}: Update
\item \textbf{Neutral Deep Research Prompt}: ``Analysis on the aftermath of SNNPR's dissolution and ongoing violence in Ethiopia. Focus on September 2024.''
\end{itemize}

\textbf{ID}: 15
\begin{itemize}
\item \textbf{Original Article Name (with link)}: \href{https://acleddata.com/report/viv-ansanm-living-together-fighting-united-alliance-reshaping-haitis-gangland}{Viv Ansanm: Living together, fighting united — the alliance reshaping Haiti's gangland}
\item \textbf{Date Publish}: 16 October 2024
\item \textbf{Article Type}: Report
\item \textbf{Neutral Deep Research Prompt}: ``How does the the Viv Ansanm alliance reshapes Haiti's gang landscape?''
\end{itemize}

\textbf{ID}: 16
\begin{itemize}
\item \textbf{Original Article Name (with link)}: \href{https://acleddata.com/report/foreign-meddling-and-fragmentation-fuel-war-sudan}{Foreign meddling and fragmentation fuel the war in Sudan}
\item \textbf{Date Publish}: 12 December 2024
\item \textbf{Article Type}: Conflict Watchlist 2025; Report
\item \textbf{Neutral Deep Research Prompt}: ``Analyze current state of war in Sudan and peace outlook, as if we are at the end of 2024 heading into 2025.''
\end{itemize}

\textbf{ID}: 17
\begin{itemize}
\item \textbf{Original Article Name (with link)}: \href{https://acleddata.com/report/mexicos-new-administration-braces-shifting-battle-lines-countrys-gang-wars}{Mexico’s new administration braces for shifting battle lines in the country’s gang wars}
\item \textbf{Date Publish}: 12 December 2024
\item \textbf{Article Type}: Conflict Watchlist 2025; Report
\item \textbf{Neutral Deep Research Prompt}: ``Research gang violence in Mexico and its outlook in 2025, as if we are at the end of 2024 heading into 2025.''
\end{itemize}

\textbf{ID}: 18
\begin{itemize}
\item \textbf{Original Article Name (with link)}: \href{https://acleddata.com/report/rwanda-defence-force-rdf-operations-abroad-signal-shift-rwandas-regional-standing}{The Rwanda Defence Force (RDF) operations abroad signal a shift in Rwanda's regional standing}
\item \textbf{Date Publish}: 27 September 2024
\item \textbf{Article Type}: Report
\item \textbf{Neutral Deep Research Prompt}: ``Explores the roles and operations of the Rwanda Defence Force in regional conflicts and their geopolitical significance.''
\end{itemize}

\textbf{ID}: 19
\begin{itemize}
\item \textbf{Original Article Name (with link)}: \href{https://acleddata.com/update/ethiopia-weekly-update-3-december-2024}{Ethiopia Weekly Update (3 December 2024)}
\item \textbf{Date Publish}: 5 December 2024
\item \textbf{Article Type}: Update
\item \textbf{Neutral Deep Research Prompt}: ``Summary of recent events in South West Ethiopia Peoples region and Oromia region, between 23 November 2024 and 3 December 2024''
\end{itemize}

\textbf{ID}: 20
\begin{itemize}
\item \textbf{Original Article Name (with link)}: \href{https://acleddata.com/report/defection-and-violence-against-civilians-sudans-al-jazirah-state-november-2024}{Defection and violence against civilians in Sudan's al-Jazirah state - November 2024}
\item \textbf{Date Publish}: 18 November 2024
\item \textbf{Article Type}: Report
\item \textbf{Neutral Deep Research Prompt}: ``Research Defection and violence against civilians in Sudan's al-Jazirah state. Focus on November 2024.''
\end{itemize}

\section{Evaluation Examples}
\subsection{Example Criteria}
\label{appendix:example_criteria}

An example of the criteria generated for the 2nd evaluation point in the dataset in shown in Table~\ref{prompt:example_criteria}, which corresponds to the ACLED article \href{https://acleddata.com/report/total-peace-paradox-colombia-petros-policy-reduced-violence-armed-groups-grew-stronger}{‘Total Peace’ paradox in Colombia: Petro's policy reduced violence, but armed groups grew stronger}.

\begin{table*}
\begin{lstlisting}[basicstyle=\ttfamily\tiny]

Name: State-armed group de-escalation
Description: Assess whether negotiations and policy shifts reduce direct clashes between security forces and armed groups, and the durability of such declines over time.

Name: Ceasefire design and oversight
Description: Evaluate the presence and effectiveness of protocols, monitoring, and enforcement mechanisms that shape ceasefire compliance and prevent opportunistic violations.

Name: Fragmentation within armed groups
Description: Analyze internal splits, leadership disputes, and factional dynamics that affect negotiation coherence, ceasefire adherence, and conflict behavior.

Name: Intergroup competition and violence
Description: Track how reduced state pressure influences armed groups' rivalries, territorial battles, and the frequency and lethality of clashes among them.

Name: Territorial expansion patterns
Description: Examine changes in geographic reach, including the number of affected localities and strategic corridors targeted for illicit economies and logistical control.

Name: Illicit economy drivers
Description: Assess how trends in drug cultivation, mining, smuggling, and extortion shape groups' finances, expansion strategies, and incentives for or against peace.

Name: Cross-border and sanctuary effects
Description: Consider how neighboring-country dynamics and safe havens influence negotiation leverage, logistics, and the persistence or resurgence of violence.

Name: Urban conflict and local actors
Description: Evaluate the role of city-based gangs and smaller groups, their ties to major outfits, and how inclusion or neglect in talks affects urban violence.

Name: Civilians' protection and harm
Description: Monitor displacement, coercion, kidnappings, and other abuses against civilians, and whether policies mitigate or unintentionally heighten these harms.

Name: Security force posture and incentives
Description: Analyze operational choices, pressure levels, and possible disincentives or collusion that shape engagement with different armed actors.

Name: Negotiation scope and sequencing
Description: Assess the government's capacity to manage multiple parallel processes, prioritize tracks, and anticipate spillovers between negotiations.

Name: Local versus national agreements
Description: Compare the effectiveness of localized, targeted deals against broad national accords in reducing violence and addressing community needs.

Name: Adaptive tactics and technologies
Description: Track shifts in armed groups' methods, such as the use of drones or curfews, and their impact on conflict intensity and state responses.

Name: Alliance fluidity and opportunism
Description: Examine how alliances and hostilities between groups change by region and over time, reflecting opportunistic calculations rather than fixed enmities.

Name: Implementation of prior accords
Description: Evaluate continuity and gaps in enforcing earlier peace agreements and reintegration efforts, and their effects on remobilization and dissidence.

Name: State presence and governance
Description: Assess whether increased governance, services, and rule of law accompany negotiations to consolidate peace and deter criminal entrenchment.

Name: Election-period volatility
Description: Monitor how political cycles affect armed group behavior, including targeted coercion and attempts to influence local authorities.

Name: Public legitimacy and trust
Description: Gauge societal support for negotiations, perceptions of fairness and accountability, and how legitimacy affects compliance and outcomes.

Name: Justice and recognition frameworks
Description: Analyze legal pathways for political recognition or judicial submission of different actor types, and how these incentives shape participation.

Name: Conflict intensity trade-offs
Description: Identify the paradox where lowering state-group confrontation can coincide with rising intergroup violence, and propose mitigating measures.
\end{lstlisting}
\caption{Example criteria generated for reference-induced criteria matching for the query in Figure~\ref{fig:figure1}.}
\label{prompt:example_criteria}
\end{table*}

\section{Prompts used}

\begin{table*}
\begin{lstlisting}[basicstyle=\ttfamily\tiny]
# instruction

You are an analytical reasoning engine that explores a relational database. Your goal is to discover surprising or meaningful insights. Your task
now is to ask new questions based on the table returned. A list of global insights is provided to you. You should NOT ask questions that are
already covered by the global insights.

Ask 1 - {{ max_questions }} specific questions to further explore relevant topics.
   - Generate 1 to {{ max_questions }} questions.
   - You are NOT required to ask all {{ max_questions }} questions.
   - Make each question self-contained and clearly scoped.
   - Follow a step-by-step process:
     1. Identify the question you are interested in.
     2. For database questions, specify the expected output format, the number of columns, and the names of those columns.
     3. Ensure the question is self-contained and clearly scoped.

For each question, also specify a "destination" to indicate where the question should be routed:
- "database": The question can be answered by querying the database (e.g., aggregations, distributions, trends, filters, correlations, rankings,
or any computation over the data).
- "internet": The question requires external context NOT available in the database (e.g., definitions of domain terms, historical context,
industry benchmarks, comparisons with external data, regulatory background, or general domain knowledge).

Most questions should be "database" - only use "internet" when the answer genuinely cannot come from the database.

Output a JSON object with:
- "chain_of_thought": your reasoning about what aspects to explore
- "questions": a list of objects, each with:
  - "question": the question text (self-contained and clearly scoped)
  - "destination": either "internet" or "database"

# input

Description of database content: {{ db_description }}

Global insights: {{ global_insights }}

Conversation history: {{ dialogue_turns }}

Topic/Question you are writing: {{ topic }}

{%
You are building evidence for the following thesis: "{{ thesis }}"

Research strategy: {{ research_strategy }}

Prioritize questions that help build, test, or refine this argument. You may also ask questions that challenge or qualify the thesis - strong analysis addresses counter-arguments.
{%
\end{lstlisting}
\caption{Exploration questions generation prompt}
\label{prompt:exploration_question_direct_gen}
\end{table*}

\begin{table*}
\begin{lstlisting}[basicstyle=\ttfamily\tiny]
# instruction

You are an analytical reasoning engine that explores a relational database. Your goal is to discover surprising or meaningful insights. Your task now is to isolate a view of the tables (with filters and/or groupbys) on which I will compute summary statistics for all columns to derive interesting insights. Generate a total of 1-{{ max_questions }} number of questions. You don't have to generate the maximum number of questions.

In each question:
- You should return a SQL of form: `SELECT * ...` based on filters and groupbys identified in the past round.
- For instance, if in the previous turn you have identified that categroy A has the most amount of population, then in this turn you might want to investiate `SELECT * FROM category = A.` Note here you should not select from multiple cateogories because summary statistics will be computed based on the result.
- First generate your reasoning and then generate the actual SQL.
- The database is PostgreSQL, so make sure to respect the syntax, such as wrapping tables in double quotes if the table name contains upper case letters.

# input

Description of database content: {{ db_description }}

Conversation history: {{ dialogue_turns }}

Topic/Question you are writing: {{ topic }}
\end{lstlisting}
\caption{Exploration direct SQL generation prompt}
\label{prompt:exploration_direct_SQL_gen}
\end{table*}

\begin{table*}
\begin{lstlisting}[basicstyle=\ttfamily\tiny]

# instruction
You are given a list of SQL responses related to the same topic. Your task is to:
1. identify any inconsistencies in the SQL predicates used and standarize any inconsistencies. For the nodes you would like to correct, issue a follow-up question with the desired SQL predicates. You can directly instruct what to modify in the SQLs. DO NOT instruct new variables not seen in the current SQL. DO NOT instruct it correct any variables.
2. Some noes will be given to you as examples. These examples will be marked with "example_node": True, and you do not need to issue a follow-up question for them.
3. make sure the SQLs reflect the conversation context presented in previous_queries. If any SQL appears to have forgotten the conversational context, issue a follow-up question to resolve it.
4. If no follow-up question is needed, set "follow_up_question": None.

Output a JSON following examples.

# input
{
    "example_node_0": {
        "query": "Show me the top 20 countries by the number of missile or artillery attacks that they have targetted by?",
        "SQL": "SELECT country, COUNT(*) AS attack_count FROM events WHERE sub_event_type IN ('Shelling/artillery/missile attack') GROUP BY country ORDER BY attack_count DESC LIMIT 20;",
        "example_node": True,
        "note": "no need to generate follow_up_question"
    },
    "query0": {
        "previous_queries": None,
        "query": "What specific regions or countries in the Middle East have seen the most significant increase in ISIS-related activities in 2025?",
        "SQL": "SELECT region, country, COUNT(*) AS event_count FROM events WHERE year = 2025 AND region = 'Middle East' AND (actor1 IN ISISactor1 OR actor2 IN ISISactor2 OR assoc_actor_2 IN ISISassoc_actor_2 OR actor1 LIKE '%
    },
    "query1": {
        "previous_queries": None,
        "query": "How have shifts in geopolitical alliances, such as Israel's potential normalization with Saudi Arabia, influenced ISIS activities in the Middle East during 2025?",
        "SQL": "SELECT * FROM event WHERE year = 2025 AND region = 'Middle East' AND (actor1 IN ISISactor1 OR actor2 IN ISISactor2 OR assoc_actor_2 IN ISISassoc_actor_2);"
    }
}
# output
{
    "query0": {
        "query": "What specific regions or countries in the Middle East have seen the most significant increase in ISIS-related activities in 2025?",
        "follow_up_question": None,
    },
    "query1": {
        "query": "How have shifts in geopolitical alliances, such as Israel's potential normalization with Saudi Arabia, influenced ISIS activities in the Middle East during 2025?",
        "follow_up_question": "Please include actor1 LIKE '%
    }
}

# input
{{ input }}

\end{lstlisting}
\caption{Query consistency module prompt}
\label{prompt:query_consistency_module}
\end{table*}

\begin{table*}
\begin{lstlisting}[basicstyle=\ttfamily\tiny]
# instruction
You are given a list of insights related to a topic. Your task is to select the most interesting and relevant insights capped at {{ max_num_insights }}.
You should NOT select similar insight twice.

The topic is: {{ topic }}

The database you are using is: {{ db_description }}

{%
## Guiding Thesis
The article being built argues: "{{ thesis }}"

Prioritize insights that are most useful for developing this argument - this includes:
- **Supporting evidence**: findings that directly build or strengthen the thesis
- **Qualifying evidence**: findings that add nuance, scope limits, or important caveats
- **Refuting evidence**: findings that challenge or contradict the thesis - strong analytical articles steel-man
counter-arguments rather than ignore them

Deprioritize insights that are entirely off-topic or redundant with others already selected.
{%

Output a JSON dict, where each key is a node_id and the value is the insight for that node_id.

# input
{{ input }}
\end{lstlisting}
\caption{Insight bank filter prompt}
\label{prompt:insight_bank_filter}
\end{table*}

\begin{table*}
\begin{lstlisting}[basicstyle=\ttfamily\tiny]
# instruction

You are a senior analyst at a world-class publication (think The Economist, Foreign Affairs, or FiveThirtyEight). You have been given a general
topic and a batch of findings produced by a preliminary data exploration agent.

Your job is NOT to describe what the data shows. Your job is to REASON about what the findings mean - to identify non-obvious patterns, causal
claims, counter-narratives, strategic implications, or surprising tensions - and to distill them into compelling, defensible thesis statements.
Each thesis should be the kind of bold, original argument that could anchor a top-tier analytical article written for a general audience.

Generate at most 3 thesis candidates.

Rules:
- Each thesis is a CONCISE TITLE - maximum 10 words. Think magazine cover line or op-ed headline, NOT a full sentence or a data summary.
- A good thesis takes a POSITION. It argues something. It should be possible to disagree with it. Avoid bland descriptive titles like "Trends in
X" or "Overview of Y."
- Do NOT embed statistics, numbers, or data citations in the thesis title.
- The thesis should capture a non-obvious, thought-provoking argument that would make an informed reader want to read the full article.
- For each thesis, provide a research_strategy: a concrete plan for how a writer should develop this argument into a full analytical article.
Specify what evidence to marshal, what comparisons to draw, what counter-arguments to address, what narrative structure to follow, and what
conclusions to build toward. This will be handed to a downstream research agent that will write the article.
- If the findings in this batch don't support 3 strong theses, output fewer. Quality over quantity.

# input

Description of database content: {{ db_description }}

Topic: {{ topic }}

Below are findings from a preliminary data exploration on this topic. Reason about what these findings reveal - the patterns, tensions, and
implications - and propose up to 3 thesis statements that could each serve as the central argument of a top-tier analytical article.

{{ context }}
\end{lstlisting}
\caption{Thesis generation prompt}
\label{prompt:thesis_generation}
\end{table*}

\begin{table*}
\begin{lstlisting}[basicstyle=\ttfamily\tiny]

# instruction

You are a senior analyst at a world-class publication (think The Economist, Foreign Affairs, or FiveThirtyEight).

You previously proposed a working thesis to guide research on a topic. Since then, a research agent has gathered
additional findings from the database. Your task is to re-examine that thesis in light of the new evidence and
decide whether to:

  1. Sharpen - narrow or deepen the original argument using new supporting evidence
  2. Pivot - shift to a better-supported or more compelling argument uncovered by the new findings
  3. Confirm - keep the thesis essentially unchanged if the evidence continues to support it strongly

Output exactly one refined thesis and the updated research strategy.

# input

Description of database content: {{ db_description }}

Topic: {{ topic }}

Current Thesis: {{ current_thesis }}

Current Research strategy: {{ current_research_strategy }}

Current findings:

{{ context }}

\end{lstlisting}
\caption{Thesis refinement prompt}
\label{prompt:thesis_refinement}
\end{table*}

\begin{table*}
\begin{lstlisting}[basicstyle=\ttfamily\tiny]

You are conducting research on a goal/topic: "{{ topic }}". The goal here is to extract previously unknown insights by exploring and observing the information in the database with the following description: {{ db_description }}.

Generate up to {{ num_questions }} questions that an investigator will be interested in. You do not need to generate all {{ num_questions }} questions if you believe you only need to ask 1-2 to get started. The questions will be used to generate search queries in the database to help answer them. The questions should be self-contained (include any specific years, months, locations, etc. instead of a reference that requires the reader to know the context) and related to the goal/topic: "{{ topic }}". Investigate any correlations as you see fit. Do not generate overly complex questions. Each question should be investigate one aspect but do not include too many subquestions inside a single question.

IMPORTANT: the questions should be independent of each other. If you believe some questions need to be answered first, only generate those questions and not others that depend on the answers to the first questions.

{%
Here is more background information on the goal/topic based on the internet: "{{ article }}".
{%

\end{lstlisting}
\caption{Initial questions generation prompt}
\label{prompt:initial_questions}
\end{table*}

\begin{table*}
\begin{lstlisting}[basicstyle=\ttfamily\tiny]

# instruction

You are planning a publication-ready analytical narrative report.
Design a single flowing report that proves the central claim through evidence.
TREE evidence is the core spine; external context should enrich but not replace it.
This is for readers, not a thesis committee.
Rules:
- The thesis is an internal organizing claim, not a standalone section heading.
- Do not plan separate "Thesis" or "Key Findings" sections.
- Use a narrative arc: opening hook -> escalation/mechanism -> geography/human stakes -> implications.
- Each section should advance the story and hand off naturally to the next.
- Every section must still materially support, test, or sharpen the thesis.
- Prefer 4 substantive sections, each capable of roughly 450-700 words.
- For each section, provide at most 3 web queries in web_queries. Queries must be specific and aimed at authoritative sources (major NGOs, official documents, major outlets), not generic search phrases.
Return JSON only with this schema:
{
  "lede_strategy": "...",
  "key_findings": ["...", "..."],
  "sections": [
    {
      "section_id": "S1",
      "heading": "...",
      "purpose": "...",
      "must_include_evidence_ids": [1,2],
      "key_points": ["..."],
      "storytelling_moves": ["..."],
      "web_queries": ["query 1", "query 2"]
    }
  ],
  "closing_strategy": "..."
}

# input
TOPIC:
{topic}

THESIS:
{thesis}

TITLE PACKAGE:
title: {{ title }}
subtitle: {{ subtitle }}
editorial_angle: {{ editorial_angle }}

CORE EVIDENCE NOTES:
{{ note_digest }}

WARMSTART CONTEXT (optional background hints):
{{ warmstart_text }}

VALID EVIDENCE IDS (must_include_evidence_ids must use only these):
{{ valid_ids }}

Create the report plan now.
\end{lstlisting}
\caption{Outline generation prompt}
\label{prompt:outline_gen}
\end{table*}

\begin{table*}
\begin{lstlisting}[basicstyle=\ttfamily\tiny]
# instruction
You are writing one section of a publication-ready analytical narrative report. Think of medias such as New York Times or the Economist.
Rules:
- TREE evidence is the core spine: prioritize it and explicitly use it.
- Supplemental web context can provide background, reactions, and scene-setting.
- Every factual claim must have inline citation(s) in [N] format.
- Use only citation numbers from ALLOWED_CITATIONS.
- Do not invent citations or facts.
- Do not repeat the section heading inside section_markdown.
- Do not open with phrases like "This section" or "This evidence". Lead with the most consequential finding.
- Move from data -> mechanism -> human stakes -> repercussions.
- Quant style: write numbers like a reporter with evidence, not a methods appendix.
  - Avoid in-line statistical jargon (e.g., "mean", "std", "p-value", "significant", "contemporaneous", "lagging", "correlation
coefficient", "r=") unless the coefficient itself is the only faithful representation of the evidence.
  - Prefer plain-language comparatives first ("about twice as high", "tracked closely", "rose sharply"), then give the exact
numbers in a second clause or sentence.
  - If you include r/lag/etc., translate immediately in plain language and avoid stacking multiple coefficients in one sentence.
  - Prefer one numeric claim per sentence; avoid dense parenthetical math.
  - Avoid meta signposting like "as later sections will detail"; use a natural bridge sentence instead.
- For each evidence citation provided, include at least one substantive use tied to that citation.
- If the web packet contains clearly relevant authoritative sources, use at least 1-2 web citations for context or external
validation. Do not force weak web sources.
- Prefer smooth prose over bullets or mini-subheadings; use internal subheadings only if truly necessary.
- End with a forward-driving sentence that naturally sets up the next section.
- Do not add any Sources/References/Citations section.
Return JSON only:
{
  "section_id": "...",
  "heading": "...",
  "section_markdown": "...",
  "used_citations": [1,2,3]
}

# input
TOPIC:
{{ topic }}

THESIS:
{{ thesis }}

REPORT TITLE:
{{ report_title }}

SECTION SPEC:
- section_id: {{ section_id }}
- heading: {{ heading }}
- purpose: {{ purpose }}
- key_points: {{ key_points }}
- storytelling_moves: {{ storytelling_moves }}

ALLOWED_CITATIONS:
{{ allowed }}

CORE TREE EVIDENCE (mandatory):
{{ core_packet }}

SUPPLEMENTAL WEB CONTEXT (optional):
{{ web_packet }}

TARGET SECTION LENGTH:
{{ target_words }} words (hard ceiling - stay under this)

Draft this section now. Keep it highly analytical, readable, and citation-grounded.

\end{lstlisting}
\caption{Section draft prompt}
\label{prompt:section_draft}
\end{table*}

\begin{table*}
\begin{lstlisting}[basicstyle=\ttfamily\tiny]

You are a precise fact-checker for data journalism reports.

You will be given a SENTENCE taken from a report and the SOURCE text that the sentence cites. Your job is to determine whether every factual claim in the
SENTENCE is supported (entailed) by the SOURCE. You will be given the context leading up to the citation.

Rules:
- Set is_entailed to true only if every factual claim in the SENTENCE can be directly verified from the SOURCE. Minor rephrasing or summarisation is fine
as long as nothing contains factual errors. For instance, fatality vs. incident count would be a factual difference.
- Set is_entailed to false if the SENTENCE adds, omits, or distorts any fact relative to the SOURCE.
- If is_entailed is false, concisely identify the issue in your output (one sentence).

DO NOT flag:
- Reasonable interpretations or paraphrases of the evidence
- Stylistic differences or summarization

# input

SENTENCE:
{{ sentence }}

SOURCE:
{{ sources }}

\end{lstlisting}
\caption{Citation grounding prompt}
\label{prompt:citation_grounding}
\end{table*}

\begin{table*}
\begin{lstlisting}[basicstyle=\ttfamily\tiny]

# instruction
You are revising one section of a publication-ready analytical narrative report.

You will be given the previous draft and a list of criticisms. Each criticism identifies a specific sentence that
makes a claim not supported by the cited evidence.

Rules:
- Fix ONLY the criticized sentences. Do not rewrite or restructure anything else.
- For each criticism, either:
  a) Rewrite the sentence to remove or qualify the unsupported claim, keeping any supported parts intact, or
  b) Remove the sentence entirely if no part of it is supportable.
- Keep all citations that remain accurate. Do not add new citations outside ALLOWED_CITATIONS.
- Do not invent new facts.
- Preserve the section's structure, flow, and all uncriticized content verbatim.
- Do not add a Sources/References/Citations section.

Return JSON only:
{
  "section_id": "...",
  "heading": "...",
  "section_markdown": "...",
  "used_citations": [1,2,3]
}

# input
TOPIC:
{{ topic }}

THESIS:
{{ thesis }}

REPORT TITLE:
{{ report_title }}

SECTION SPEC:
- section_id: {{ section_id }}
- heading: {{ heading }}
- purpose: {{ purpose }}
- key_points: {{ key_points }}
- storytelling_moves: {{ storytelling_moves }}

ALLOWED_CITATIONS:
{{ allowed }}

CORE TREE EVIDENCE (mandatory):
{{ core_packet }}

SUPPLEMENTAL WEB CONTEXT (optional):
{{ web_packet }}

PREVIOUS DRAFT:
{{ previous_draft }}

CRITICISMS:
{{ criticisms }}

Revise the section now. Change only what the criticisms require.

---
Key differences from the draft prompt:
- Instructions replace "write from scratch" with "fix only criticized sentences" - prevents the model from rewinding  
and redrafting the whole section                                                                                    
- The criticisms variable is the JSON list of {"original_sentence", "criticism"} dicts, so the model can match exactly
 what to fix                                                                                                          
- No word count / paragraph count rules - output length is determined by what was kept                                
- previous_draft, criticisms added as new variables; everything else identical so the model still has full evidence 
context

\end{lstlisting}
\caption{Section revision prompt}
\label{prompt:section_revision}
\end{table*}

\begin{table*}
\begin{lstlisting}[basicstyle=\ttfamily\tiny]

# instruction
You are a senior editor polishing a near-final publication-ready report draft. Think of medias such as New York Times or the Economist.
Rules:
- Preserve and improve analytical flow, and add smooth transitions between sections.
- Preserve or expand substance; do not compress the draft into a summary.
- Keep all existing valid citations; do not invent new citation numbers.
- Use only citation numbers listed in ALLOWED_CITATIONS.
- Keep markdown headings and publication-ready prose.
- Do NOT add Sources/References/Citations section (it will be appended programmatically).
- The final polished report body must not exceed {{ target_total_words }} words (excluding the sources appendix). Cut ruthlessly for concision while preserving every cited claim.
- You should include a conclusion section at the end.
- Do not explictly include a "thesis" block
Return JSON only:
{
  "report_markdown": "..."
}

# input
TOPIC:
{{ topic }}

THESIS:
{{ thesis }}

TITLE:
{{ title }}
{{ subtitle }}

PLAN (for structural intent):
{{ plan_json }}

ALLOWED_CITATIONS:
{{ allowed_citations }}

DRAFT REPORT:
{{ draft_markdown }}
\end{lstlisting}
\caption{Final polish prompt}
\label{prompt:final_polish}
\end{table*}

\begin{table*}
\begin{lstlisting}[basicstyle=\ttfamily\tiny]

# instruction

You are an expert analyst. Given a reference research article, extract a list of evaluation criteria describing
what analytical points a good report on this topic should cover. Focus on general trends and patterns - do not
reference specific numbers, dates, or proper nouns that would make the criteria too narrow.

Read the reference article carefully. Identify the key analytical points it makes - the insights, trends, and
conclusions that a thorough report on this topic should include.

For each criterion:
1. Give it a short **name** (3-6 words)
2. Write a **description** of the general trend or pattern to look for (1-2 sentences, no specific numbers or
dates needed but include e.g. the general trend)

Return as a JSON object with a "criteria" array, each item having "name" and "description" fields.

# input

## Research Task
{{task_prompt}}

## Reference Article
{{reference_article}}

\end{lstlisting}
\caption{Reference-induced criteria generation prompt}
\label{prompt:criteria-matching-gen}
\end{table*}

\begin{table*}
\begin{lstlisting}[basicstyle=\ttfamily\tiny]
# instruction

You are an expert evaluator of analytical research articles. Given a set of evaluation criteria and a generated
article, grade how well the article addresses each criterion.

For each criterion, assess how well the generated article addresses it on a 0.0-1.0 scale:
- **1.0** - Fully addresses: the article clearly covers this analytical point
- **0.75** - Mostly addresses: covered but with gaps or insufficient depth
- **0.5** - Partially addresses: touches on it but misses key aspects
- **0.25** - Barely addresses: only a brief or tangential mention
- **0.0** - Not addressed: completely absent from the article

Return as a JSON object with fields: criterion_scores (array of {name, score, explanation}).

# input

{%
{{ score_reminder }}
{%
## Research Task
{{task_prompt}}

## Evaluation Criteria
{{criteria}}

## Generated Article
{{generated_article}}
\end{lstlisting}
\caption{Reference-induced criteria grading prompt}
\label{prompt:criteria-matching-grade}
\end{table*}

\begin{table*}
\begin{lstlisting}[basicstyle=\ttfamily\tiny]
Given an input article, your task is to break down the insights in the article into itemized points. Each insight should be self-contained.

Input article: {{ article }}
\end{lstlisting}
\caption{Atomic breakdown prompt}
\label{prompt:atomic-breakdown}
\end{table*}

\begin{table*}
\begin{lstlisting}[basicstyle=\ttfamily\tiny]
You are an expert analyst evaluating whether a piece of evidence from a generated article is derived from ACLED (Armed Conflict
Location & Event Data) data.

Article topic: {{article_topic}}

Evidence to classify: {{evidence}}

ACLED data includes:
- Conflict event counts, incident reports, and event descriptions
- Violence against civilians statistics
- Battle-related data (battles, explosions/remote violence, riots, protests)
- Fatality counts and casualty figures from conflict events
- Geographic conflict data (locations of events, subnational breakdowns)
- Conflict trend analysis and temporal patterns derived from event data
- Armed group activity and actor-level data
- Conflict index scores (e.g., ACLED Conflict Index)
- Data explicitly attributed to ACLED or its datasets

NOT ACLED data:
- General geopolitical analysis or commentary not tied to specific event data
- Economic indicators (GDP, inflation, trade figures)
- Humanitarian statistics from UN agencies (UNHCR refugee counts, OCHA displacement figures) unless explicitly tied to ACLED
- Demographic or census data
- Policy statements, diplomatic actions, or government declarations
- Media reports or journalistic analysis without specific conflict event data
- Academic or think-tank analysis not grounded in ACLED event data
- Data from other conflict databases (e.g., UCDP, GTD, IISS) unless attributed to ACLED
\end{lstlisting}
\caption{Insight attribution prompt}
\label{prompt:insight-attribution}
\end{table*}

\begin{table*}
\begin{lstlisting}[basicstyle=\ttfamily\tiny]
Below is an instruction that describes a task. Write a response that appropriately completes the request.

### Instruction:
Provided Answer:
{answer}

Ground Truth Answer:
{gt_answer}

Follow these instructions when writing your response:
* On a scale of 1-10, provide a numerical rating for how close the provided answer is to the ground truth answer, with 10 denoting that the provided answer is the same as ground truth answer.
* Your response should contain only the numerical rating. DONOT include anything else like the provided answer, the ground truth answer, or an explanation of your rating scale in your response.
* Wrap your numerical rating inside <rating></rating> tags.
* Check very carefully before answering.
* Follow the output format as shown in the example below:
Example response:
<rating>7</rating>

### Response:
\end{lstlisting}
\caption{InsightBench eval prompt}
\label{prompt:insightbench-eval}
\end{table*}

\begin{table*}
\begin{lstlisting}[basicstyle=\ttfamily\tiny]

# instruction
Your task is to write a **{{ database_type }}** query to answer the given question. Follow a step-by-step process:

- User question is contextual. If needed, the current date is {{ curr_date }}
- Start by constructing simple fragments of the **{{ database_type }}** query.
- Execute each fragment to verify its correctness. Adjust as needed based on your the observations.
- Confirm all your assumptions about the structure of the database before proceeding.
- Do NOT repeat the same action, as the results will be the same.
- You will always be shown with a sample of database results. If user is asking for all entries, entire results will be displayed in a seperate module.
- Output one final SQL at the end that contains all results. You can only output one SQL query (not multiple ";" separated queries).
- For stats/visualizations, you should call the execute_python_from_sql action AFTER you have determined the final SQL query.
- Only the latest plot will be shown to you.

Form exactly one "Thought" and perform exactly one "Action", then wait for the "Observation".

Possible actions are:

- get_tables(): Retrieves all the tables with a corresponding short description. **You should use this action to get the available tables**
- retrieve_tables_details([table_names]): Retrieve more details about table(s). The argument should be a list of table names as strigns. // Example: retrieve_tables_details(["table1"]) or retrieve_tables_details(["table1", "table2"])
- execute_sql(sql): Runs a SQL query and returns results. 
- execute_python_from_sql(sql, python_code): Executes a Python code based on the results of the SQL query. The argument should be a Python tuple containing the SQL query and the Python code. Your python code can reference the results of the SQL query using the `sql_results` variable, which will be passed in as a pandas dataframe. The two codes should be wrapped in string quotes(") // Example: execute_python_from_sql("SELECT * FROM table1", "print(sql_results)")
- stop(): Marks the last executed SQL query as the final answer and ends the process. You can directly use this if user is not engaged in database-related conversation but instead just chit-chatting.

# input
Prior turn contexts:
--
{%
User Question: {{ conversation_history[i]["question"] }}

Action history:
{%
{{ conversation_history[i]["action_history"][j] }}
{%

Agent Response: {{ conversation_history[i]["response"] }}
--
{%
Current-turn User Question: {{ question }}

{%
Action history:
{%

{{ action_history[i] }}
{%
{%

Output one "Thought" and one "Action":

\end{lstlisting}
\caption{Executor main prompt}
\label{prompt:executor-main}
\end{table*}

\begin{table*}
\begin{lstlisting}[basicstyle=\ttfamily\tiny]
You are given a list of data insights derived from a dataset analysis.
Write a concise, coherent paragraph (3-5 sentences) that summarizes the key findings.
Focus on the most important patterns and avoid repetition.\n\n
Insights:\n{{insights}}
\end{lstlisting}
\caption{InsightBench summary prompt}
\label{prompt:summary_prompt}
\end{table*}

\FloatBarrier
\begin{lstlisting}[caption={Newsworthiness rubric used in human evaluation.},label={lst:newsworthiness-rubric},basicstyle=\ttfamily\small,breaklines=true,frame=single]
rubric_items = [
  {
    id: "NW1",
    name: "Public impact / magnitude",
    description: "Rate how large, serious, and consequential the report's central investigative angle appears to be. Consider whether the story points to substantial harms, large numbers of affected people, major resource implications, broad institutional consequences, or systemic rather than isolated effects. A high score should reflect a story whose consequences would matter at significant scale if true; a low score should reflect a story whose consequences remain narrow, local, or limited.",
    likert_scale: {
      1: "Very weak impact: The story's consequences appear minor, highly localized, or limited to a very small set of actors. The stakes seem too small to make the angle strongly newsworthy.",
      2: "Weak impact: The story shows some real consequences, but they remain fairly bounded in scale or seriousness. The issue matters, but not at a level that strongly elevates editorial priority.",
      3: "Adequate impact: The story presents meaningful and credible consequences with clear importance, but the scale is moderate rather than especially large or far-reaching.",
      4: "Strong impact: The story shows substantial consequences, such as broad public effects, serious harms, significant institutional implications, or clearly systemic importance.",
      5: "Exceptional impact: The story points to consequences of outstanding scale or severity, such as very large populations affected, major systemic failures, or effects that would be difficult for an editor to ignore."
    }
  },
  {
    id: "NW2",
    name: "Public-interest / accountability value",
    description: "Rate how strongly the report serves a watchdog or accountability function. Consider whether the report reveals information the public genuinely needs to know because it bears on governance, public safety, abuse of power, hidden risks, rights, corruption, institutional failure, or other matters of civic importance. A high score should reflect a story with clear public-interest necessity; a low score should reflect a story that may be interesting but has weak accountability or civic value.",
    likert_scale: {
      1: "Very weak public-interest value: The report offers little meaningful watchdog, civic, or accountability function. The public need to know this is weak.",
      2: "Weak public-interest value: The report has some relevance to public concerns, but its accountability significance is modest or secondary.",
      3: "Adequate public-interest value: The report presents a clear and legitimate public-interest rationale, though the accountability stakes are moderate rather than especially strong.",
      4: "Strong public-interest value: The report substantially informs public understanding of institutional behavior, public risk, or misuse of power, making it clearly valuable as investigative journalism.",
      5: "Exceptional public-interest value: The report appears to reveal information the public urgently or decisively needs in order to understand, scrutinize, or respond to highly consequential conduct or failure."
    }
  },
  {
    id: "NW3",
    name: "Power / institutional significance",
    description: "Rate how centrally the report engages with powerful actors, institutions, or systems. Consider whether the story concerns governments, militaries, regulators, large corporations, dominant organizations, or other actors whose decisions and conduct have outsized effects. A high score should reflect a report that meaningfully illuminates the behavior or failures of major power centers; a low score should reflect a report focused mainly on marginal, private, or low-significance actors.",
    likert_scale: {
      1: "Very weak institutional significance: The story mainly concerns minor private actors or low-level incidents, with little connection to meaningful centers of power.",
      2: "Weak institutional significance: Some institutions or organizations are mentioned, but they are not central to why the story matters.",
      3: "Adequate institutional significance: Relevant institutions or influential actors are materially involved, though the story is not primarily driven by major power structures.",
      4: "Strong institutional significance: Major institutions, governing actors, or system-level forces are central to the story, and their involvement clearly raises its editorial importance.",
      5: "Exceptional institutional significance: The story directly concerns highly consequential power structures or top-tier institutions, and understanding their conduct is fundamental to the story's significance."
    }
  },
  {
    id: "NW4",
    name: "Conflict / contestation / stakes",
    description: "Rate how strongly the report reveals meaningful conflict, contradiction, or unresolved tension. This can include armed conflict, political contestation, legal disputes, contradictory evidence, competing institutional claims, or clashes of interest. A high score should reflect a story in which the tension is substantial and central to why the angle matters; a low score should reflect a story that is largely descriptive, uncontested, or low-stakes.",
    likert_scale: {
      1: "Very weak conflict: The report contains little real dispute, contradiction, or tension. The angle feels informational but not strongly contested.",
      2: "Weak conflict: Some disagreement or tension is present, but it seems limited in significance or peripheral to the main angle.",
      3: "Adequate conflict: The report identifies a clear and relevant dispute, contradiction, or unresolved tension that contributes to its news value.",
      4: "Strong conflict: The report centers on a substantial clash, contradiction, or high-stakes dispute that clearly heightens the importance of the story.",
      5: "Exceptional conflict: The report reveals an intense, deeply consequential, or highly unstable conflict or contradiction that strongly drives urgency and editorial priority."
    }
  },
  {
    id: "NW5",
    name: "Originality / newness / exclusivity",
    description: "Rate how much genuinely new journalistic value the report appears to contribute. Consider whether it presents new findings, non-obvious synthesis, unusual access, fresh evidence, or a distinctive investigative angle rather than reiterating familiar background. A high score should reflect a report that materially advances understanding of the issue; a low score should reflect one that feels derivative, expected, or only superficially fresh.",
    likert_scale: {
      1: "Very weak originality: The report feels largely familiar, derivative, or unsurprising, with little apparent addition to existing understanding.",
      2: "Weak originality: The report has limited freshness, mostly in framing or presentation rather than in the substance of the angle.",
      3: "Adequate originality: The report offers some distinctiveness, fresh synthesis, or a meaningful but not major contribution beyond routine coverage.",
      4: "Strong originality: The report clearly adds new journalistic value through a fresh angle, notable finding, unusual access, or meaningful interpretive advance.",
      5: "Exceptional originality: The report feels genuinely revelatory, highly distinctive, or scoop-like, substantially advancing what an informed editor or reader would understand about the issue."
    }
  },
  {
    id: "NW6",
    name: "Human consequence / concreteness",
    description: "Rate how clearly the report makes the stakes concrete in human terms. Consider whether it shows how the issue affects people, communities, or lived conditions in specific and tangible ways rather than remaining abstract, purely technical, or purely institutional. A high score should reflect a report that makes the consequences unmistakably real without relying on empty emotionality; a low score should reflect a report whose stakes remain vague or abstract.",
    likert_scale: {
      1: "Very weak concreteness: The report remains abstract, technical, or detached, and the real-world consequences are difficult to grasp.",
      2: "Weak concreteness: Some human effects are mentioned, but they are thin, generic, or weakly connected to the central angle.",
      3: "Adequate concreteness: The report gives a workable and reasonably clear sense of how the issue affects people or communities.",
      4: "Strong concreteness: The report clearly demonstrates specific lived consequences that make the stakes tangible and strengthen the investigative angle.",
      5: "Exceptional concreteness: The report makes the human consequences especially vivid, specific, and consequential, so that the stakes feel unmistakably real and editorially compelling."
    }
  },
  {
    id: "NW7",
    name: "Continuing value / follow-up potential",
    description: "Rate whether the report appears to open a durable line of coverage rather than a one-off curiosity. Consider whether the issue seems ongoing, unresolved, likely to generate response, likely to develop further, or likely to support additional reporting and editorial attention. A high score should reflect a story with strong continuing value; a low score should reflect one whose relevance appears short-lived or exhausted by a single report.",
    likert_scale: {
      1: "Very weak continuing value: The story appears mostly one-off, with little indication that it would sustain follow-up coverage or further developments.",
      2: "Weak continuing value: The story may invite limited additional attention, but its follow-up potential appears modest.",
      3: "Adequate continuing value: The story has plausible follow-up potential and could justify some continued reporting, though not necessarily as a major editorial thread.",
      4: "Strong continuing value: The report points to a clearly durable issue with substantial room for follow-up reporting, response, or future developments.",
      5: "Exceptional continuing value: The report appears to open a highly sustainable and consequential line of coverage that could drive substantial future reporting and editorial attention."
    }
  }
]
\end{lstlisting}